%% file: main.tex
\documentclass[article,10pt]{wlscirep}
\usepackage[utf8]{inputenc}
\usepackage[T1]{fontenc}
\usepackage{todonotes}
\usepackage{amsmath}
\usepackage{graphicx}
\usepackage{authblk}
\usepackage{xcolor}
\usepackage{colortbl}
\usepackage{todonotes}
\usepackage[colorlinks=true, allcolors=blue]{hyperref}
\usepackage{fontspec}
\usepackage{multirow}
\usepackage{wrapfig}
\usepackage{polyglossia}
\usepackage{url}
\usepackage{subfig}
\RequirePackage{filecontents}
\usepackage{lineno}

\usepackage{float}
\usepackage{algorithm}
\usepackage{algpseudocode}
\usepackage{fontspec}
\usepackage{multirow}
\usepackage{polyglossia}
\usepackage{subcaption}
\usepackage{natbib}
\usepackage[most]{tcolorbox}

\definecolor{maroon}{cmyk}{0,0.87,0.68,0.32}

\title{Latent Performance Profiling of Large Language Models}

\author[1,2]{Tanmoy Chakraborty\thanks{Corresponding author: tanchak@iitd.ac.in. First two authors contributed equally. The remaining authors are listed in alphabetical order by surname.}}
\author[1]{Ayan Sengupta}
\author[3]{Suparna Bhattacharya}
\author[4]{Partha Pratim Chakrabarti}
\author[5,9]{Amlan Chakrabarti}
\author[6]{Supratik Chakraborty}
\author[7]{Partha Pratim Das}
\author[7]{Lipika Dey}
\author[8]{Richa Singh}
\author[8]{Mayank Vatsa}

\affil[1]{Department of Electrical Engineering, Indian Institute of Technology Delhi, New Delhi, India}
\affil[2]{Yardi School of Artificial Intelligence, Indian Institute of Technology Delhi, New Delhi, India}
\affil[3]{Hewlett Packard Enterprise, Bangalore, India}
\affil[4]{Department of Computer Science \& Engineering, Indian Institute of Technology Kharagpur, India}
\affil[5]{A.K.Choudhury School of Information Technology, University of Calcutta, India}
\affil[6]{Department of Computer Science \& Engineering, Indian Institute of Technology Bombay, India}
\affil[7]{Department of Computer Science, Ashoka University, Sonipat, India}
\affil[8]{Department of Computer Science \& Engineering, Indian Institute of Technology Jodhpur, India}
\affil[9]{Department of Artificial Intelligence, Indian Institute of Technology Kharagpur, India}

\makeatletter
\renewcommand\@biblabel[1]{}

\makeatother

\begin{abstract}
Large language models (LLMs) frequently achieve impressive scores on standardized benchmarks, yet accuracy alone offers a limited view of their capabilities. Evaluating open-source LLMs on leaderboards faces persistent issues such as data contamination, a narrow task scope, and poor alignment with real-world reliability. Benchmark-based evaluations such as MMLU-Pro, BBH, or IFEval primarily capture \textit{what} a model outputs on fixed test sets, not \textit{how} it processes information, calibrates uncertainty, or structures internal knowledge. In this article, we advocate for a shift from benchmark-centric evaluation toward a complementary, \textit{state-centered intrinsic assessment} of LLMs. To this end, we introduce \textbf{Latent Performance Profiling (LPP)} --- a framework that derives task-agnostic diagnostics from hidden activations and output distributions. LPP defines a set of scalar metrics on a model's latent representations and dynamics, revealing traits that enable interpretable comparisons and uncover hidden vulnerabilities. Unlike static accuracy scores, LPP provides stable, architecture-sensitive signatures across models of similar size. With extensive empirical analyses across eight LLMs, spanning a size range of 0.5B-14B, we demonstrate that models with similar benchmark scores can exhibit contrasting latent profiles, such as differences in entropy or compressibility. Guided by these insights, we design synthetic probes for uncertainty and symbolic reasoning that align with intrinsic metrics while decoupling from leaderboard bias. We recommend reporting LPP alongside benchmarks to provide a deeper, interpretable understanding of model behavior, enabling more reliable model selection, safety assessment, and evaluation beyond surface-level accuracy.

\end{abstract}

\begin{document}

\flushbottom
\maketitle

As large language models (LLMs) continue to dominate public leaderboards~\citep{chang2023survey,liang2023helm}, concerns are growing that such leaderboard-driven evaluations conflate genuine reasoning ability with brittle, surface-level pattern recognition~\citep{banerjee2024vulnerability, fodor2025linegoesupinherent, navarro2025brittle}.
Recent open-source models, such as LLaMA~\citep{dubey2024llama}, Qwen~\citep{qwen2025qwen2}, Mixtral~\citep{jiang2024mixtral}, and Mistral~\citep{jiang2023mistral7b}, approach or match the performance of proprietary systems across various domains, including knowledge tasks, reasoning, and code generation, and LLMs are increasingly explored for scientific, medical, and educational applications~\citep{boiko2023autonomous,miret2025enabling,xin2025towards,agrawal2025evaluation,milano2023large}. However, key questions remain about how these models internally process information and whether existing benchmark accuracy meaningfully reflects underlying capability or reliability. Most leaderboards, such as those built on MMLU~\citep{hendrycks2021mmlu}, GSM8K~\citep{cobbe2021training}, and HumanEval~\citep{chen2021evaluating}, report performance on fixed, task-specific test sets. To expand coverage, newer evaluations such as MMLU-Pro~\citep{wang2024mmluprorobustchallengingmultitask}, IFEval~\citep{zhou2023ifeval}, and BIG-bench~\citep{srivastava2022bigbench} target multi-step reasoning, instruction following, and social cognition. While these advances improve task coverage, they remain extrinsic and evaluate only final predictions, not the model's internal decision-making process.

This output-only focus limits interpretability and reliability. Leaderboard gains are often inflated by data contamination~\citep{sainz2023trouble, zhao2024mmlu}, prompt engineering, or overfitting to benchmark-specific patterns~\citep{banerjee2024vulnerability}. Furthermore, Goodhart's Law suggests that once benchmarks become optimization targets, they lose diagnostic utility~\citep{fodor2025linegoesupinherent}. Despite high scores, models continue to hallucinate~\citep{farquhar2024detecting,bang2025hallulens}, fail under distribution shift, and exhibit brittle reasoning~\citep{zhou2024larger}. Core competencies such as uncertainty calibration, robust compositionality, or structured reasoning remain largely unmeasured.

These limitations are further highlighted by inconsistencies among models of comparable scale and complexity. A 70B-parameter model may excel at code generation, whereas another of comparable size may perform better in dialogue or reasoning. Public leaderboards collapse such nuanced differences into a single composite score, obscuring meaningful variation. Although scaling trends suggest that performance generally improves with model size~\citep{kaplan2020scaling,hernandez2021scaling}, this relationship can be misleading~\citep{zhou2024larger}. As~\citet{fodor2025linegoesupinherent} argued, progress on narrow benchmarks does not necessarily imply broader generalization. Empirical analyses reveal brittle internal representations, where trivial perturbations in input can derail outputs~\citep{haller2025llm}. Moreover, genuinely emergent abilities tend to appear only beyond specific scaling thresholds, below which performance may stagnate or even regress. Consequently, benchmark metrics often provide an incomplete and sometimes distorted view of model capabilities.

We propose \textbf{Latent Performance Profiling (LPP)} as a complementary framework for evaluating LLMs. Unlike traditional benchmarks, LPP probes \textit{how} a model processes information by measuring internal dynamics --- specifically, predictive uncertainty, representational compression, and activation diversity. Building on entropy-based uncertainty measures~\citep{kuhn2023semantic,kadavath2022language} and spectral measures of representations~\citep{wei2024large,jha2025spectral,skean2025layer}, we use three task-agnostic metrics: minimum next-token entropy (confidence), effective rank of hidden-state covariances (dimensionality), and participation ratio (variance distribution). These metrics, computed on unlabeled prompts, form compact latent profiles whose model rankings are \color{black}broadly stable across sample sizes and, for PR and ER, across calibration datasets. \color{black} Crucially, LPP reveals performance-relevant traits invisible to benchmark accuracy.

Our empirical analyses span open-source LLMs with parameter counts ranging from 0.5B to 14B, including the LLaMA, Qwen, and Mistral families. We find that models with similar benchmark accuracy (e.g., on IFEval or BBH) diverge widely in LPP space, exhibiting distinct uncertainty floors and activation geometries. Our results show how two models --- LLaMA-3B and Qwen-3B --- achieve comparable accuracy but differ in entropy and representational spread, indicating fundamentally different reasoning strategies. Layerwise analysis further reveals a mid-layer bottleneck in PR and ER in all eight models, with an additional early-layer peak (an hourglass profile) in the Qwen family, in line with earlier layer-wise analyses of transformer representations~\citep{valeriani2023the}.

\begin{figure}[t!]
    \centering
    \includegraphics[width=\linewidth]{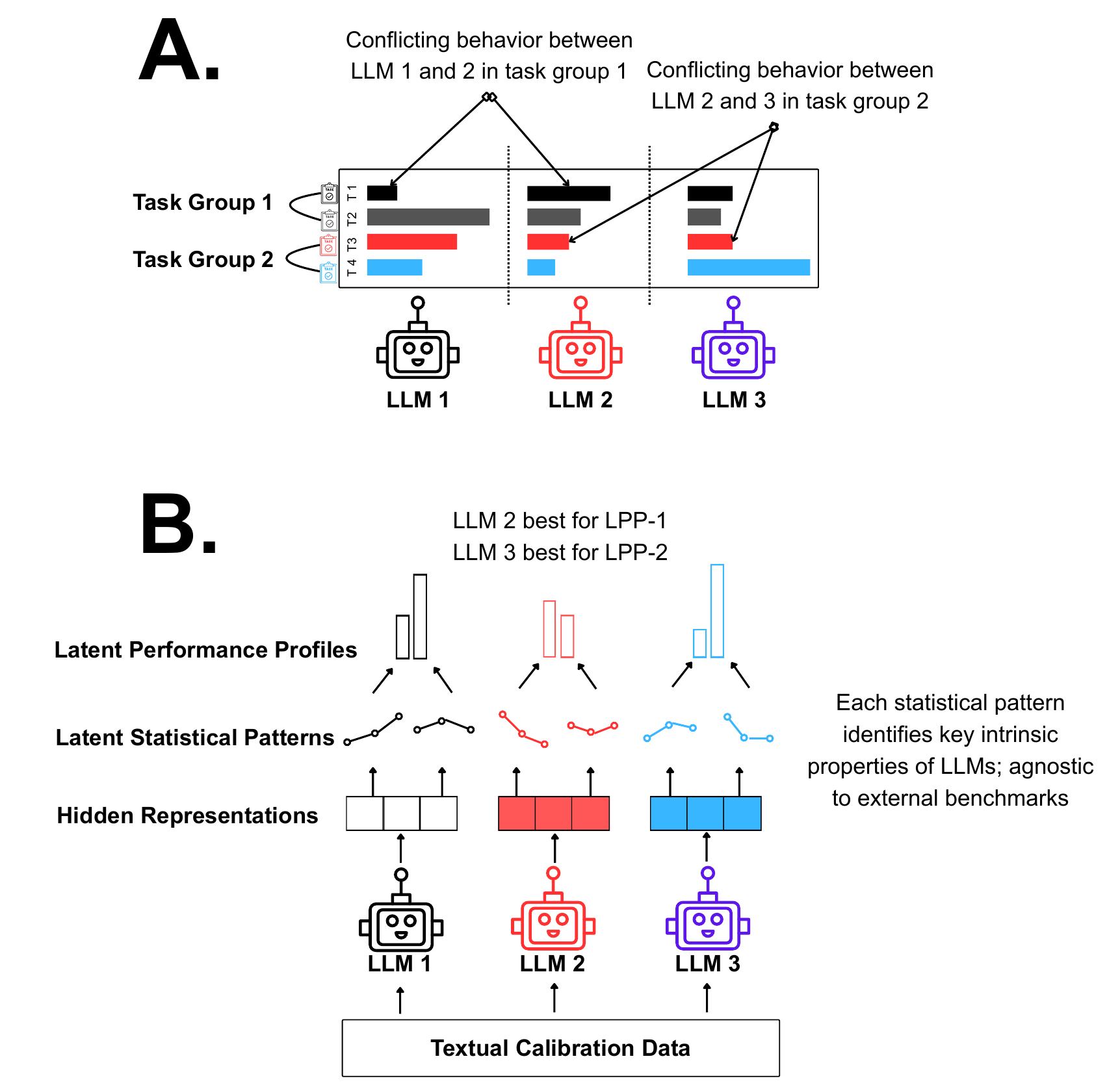}
    \caption{\textbf{Our proposed framework --- Latent Performance Profiling (LPP)}, aims to uncover differences in LLM behavior that extrinsic benchmarks may miss. While traditional evaluations might suggest similar capabilities across models, LPP probes their \textit{internal activations and representations} to reveal deeper distinctions. \textbf{(A)} We highlight instances where three LLMs are evaluated on four downstream tasks, across two task groups. LLM 1 and 2 demonstrate conflicting behavior in task group 1, wherein one model achieves higher performance in one task, while another excels in the other. These behaviors (empirically highlighted in Figure~\ref{fig:motivation_result}) are \textit{unreliable} for a practitioner selecting a model for a similar new task. \textbf{(B)} LPP-driven benchmarks, grounded in intrinsic statistical patterns, offer more reliable signals. For instance, a model with the lowest entropy may be better suited for tasks that require minimizing hallucination risk.} \label{fig:latent_motivation}
\end{figure}

To demonstrate the practical value of LPP, we introduce two synthetic diagnostic tasks --- Ambiguous Reasoning (AR) and Symbolic Pattern Completion (SPC) --- designed to stress-test entropy calibration and representational compression. Performance on these tasks correlates strongly with the corresponding LPP metrics and separates models more clearly than traditional benchmarks.
Figure~\ref{fig:latent_motivation} illustrates how LPP can reveal variation among models with superficially similar benchmark scores. For example, one model may display low predictive entropy, a common uncertainty signal~\citep{farquhar2024detecting,sharma2025think}, and compact internal states, while another may exhibit high entropy and redundant internal states, suggesting fragility despite comparable accuracy. Representation-level properties of this kind can matter for network behavior; for instance, the effective rank of hidden representations has been linked to loss of plasticity in continual learning~\citep{dohare2024loss}. In this work, we formalize such latent behaviors through a set of interpretable LPP metrics and provide a detailed empirical analysis of how these internal patterns vary across model families, sizes, and inputs. While our proposed metrics --- entropy, participation ratio, and effective rank --- are not exhaustive, they offer a principled starting point for state-based evaluation. We view LPP as a ``litmus test'' that probes essential internal properties of LLMs, offering complementary insight to traditional benchmarks and motivating the development of more comprehensive latent evaluation frameworks in the future.


\section*{Results}

\subsection*{Benchmark performance inconsistencies}
LLMs are primarily evaluated on public leaderboards such as MMLU~\citep{hendrycks2021mmlu}, GSM8K~\citep{cobbe2021training}, and HumanEval~\citep{chen2021evaluating}, which track general knowledge, mathematical reasoning, and coding ability, respectively. While these benchmarks indicate steady progress, they often conceal structural limitations, including data contamination~\citep{sainz2023trouble} and overfitting driven by Goodhart's Law~\citep{banerjee2024vulnerability}. Newer suites broaden task coverage but still offer limited insight into a model's reasoning processes or robustness. Moreover, models frequently rank differently across leaderboards~\citep{zhao2024mmlu}, leaving practitioners uncertain which model to deploy for a given application.

To quantify these inconsistencies, we analyze published benchmark scores for eight open-source models --- LLaMA-3B, LLaMA-8B, Qwen-0.5B, Qwen-1.5B, Qwen-3B, Qwen-7B, Qwen-14B, and Mistral-7B, on three representative suites: MMLU-Pro, IFEval, and BIG-Bench Hard (BBH)~\citep{suzgun2023challenging}, covering knowledge, instruction-following, and multi-step reasoning. As expected, larger models generally achieve higher scores, consistent with established scaling laws~\citep{hernandez2021scaling,kaplan2020scaling}. However, size alone is an unreliable predictor of leaderboard performance. Qwen-3B outperforms Mistral-7B on IFEval (64.7\% versus 55.0\%). Likewise, Mistral-7B surpasses Qwen-1.5B on IFEval and BBH (55.0\% versus 44.8\%, and 22.9\% versus 19.8\%, respectively), but slightly trails on MMLU-Pro (19.1\% versus 20.0\%). Even models with nearly identical architectures yield inconsistent rankings: LLaMA-8B scores only 0.2 percentage points higher than LLaMA-3B on IFEval, despite being over twice the size. At higher performance tiers, recent model families cluster tightly in score (for instance, LLaMA-3 and Qwen2.5 are nearly indistinguishable on these benchmarks; Figure~\ref{fig:motivation_result}A), making it difficult to determine which model is truly superior for real-world tasks.

These findings highlight the limitations of leaderboard-centric evaluation. Benchmarks are indispensable for measuring task success, but they do not reveal how models arrive at answers or why similar accuracy can stem from divergent reasoning pathways. High aggregate scores often hide failure modes such as miscalibration, brittle intermediate representations, or overly high-dimensional activations that are sensitive to context shifts. This motivates the need for complementary diagnostics that move beyond surface metrics to characterize internal model behavior, latent risk, and reliability. 

\begin{figure}[!t]
    \centering
    \includegraphics[width=\linewidth]{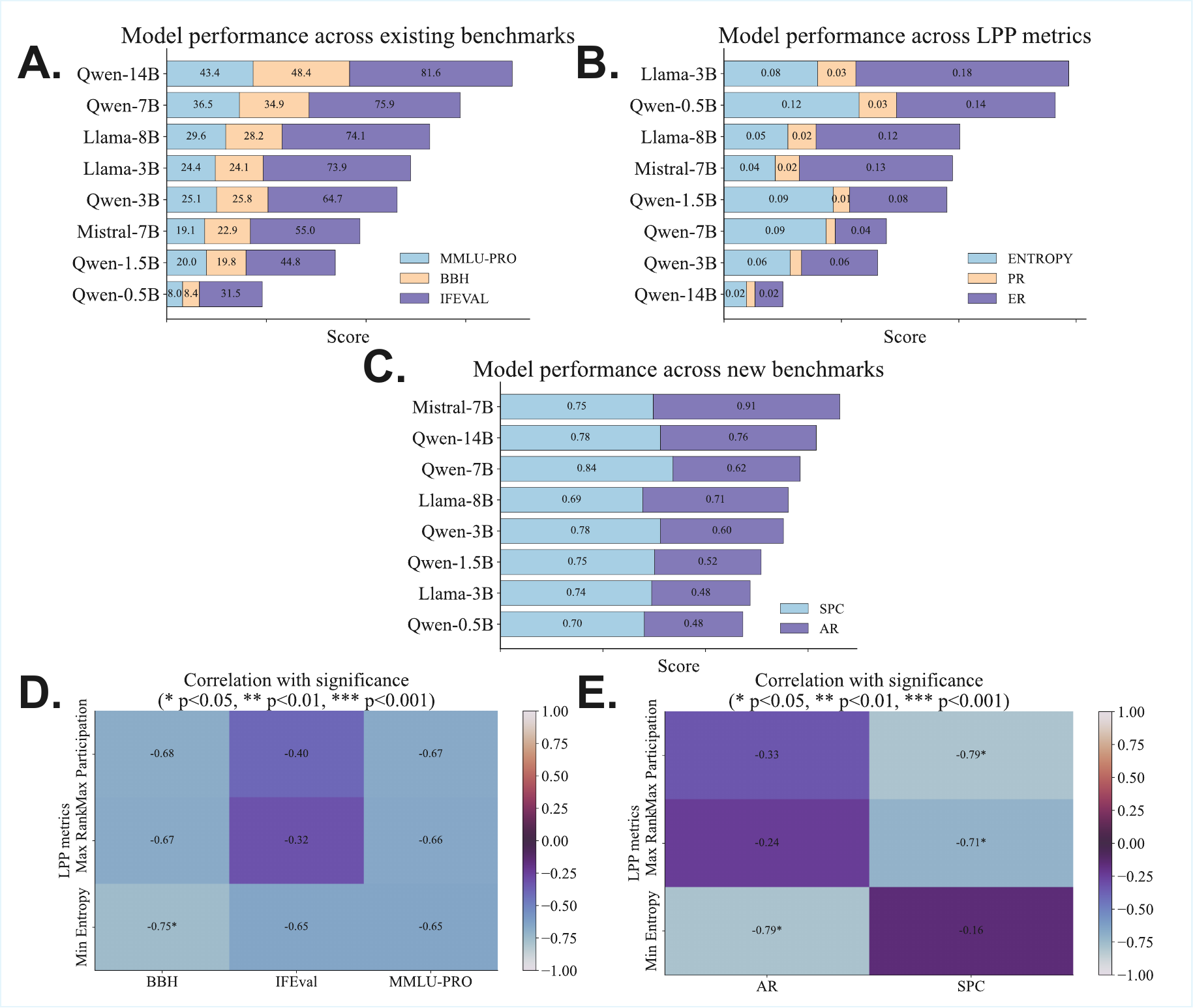}
    \caption{\textbf{Extrinsic benchmark performance, latent profiling, and correlation with new LPP-driven tasks.} This figure summarizes the core empirical findings of our study. \textbf{(A)} Comparison of model performance across three widely used extrinsic benchmarks (MMLU-Pro, BBH, and IFEval), revealing size-dependent trends and inconsistency across tasks --- larger models do not uniformly outperform smaller ones. \textbf{(B)} LPP metrics --- Entropy, PR, and ER values for different LLMs. All profiles use the same schedule (prefixes of 20--150 tokens; see Methods). \textbf{(C)} Performance on two synthetic, LPP-inspired benchmarks --- Ambiguous Reasoning (AR) and Symbolic Pattern Completion (SPC), designed to target specific internal capacities (entropy modulation and representation compression, respectively). Unlike standard tasks, these LPP-driven benchmarks show clearer structure across models. \textbf{(D)} Pearson correlation between these benchmark scores and intrinsic capacity metrics from Latent Performance Profiling (minimum entropy, maximum effective rank, and maximum participation ratio). Correlations are uniformly negative but mostly non-significant with eight models. \textbf{(E)} Performance on these synthetic tasks correlates significantly with the corresponding latent metrics, particularly entropy, thereby validating their interpretability and diagnostic alignment. Both (D) and (E) use an unadjusted two-sided Pearson correlation test for statistical significance. Together, we show how LPP metrics provide a complementary lens for explaining heterogeneous model behavior, resolving benchmark ambiguities, and supporting the principled design of future evaluation tasks.}
\label{fig:results}
\end{figure}

\begin{figure}[t!]
    \centering
    \includegraphics[width=1.05\linewidth]{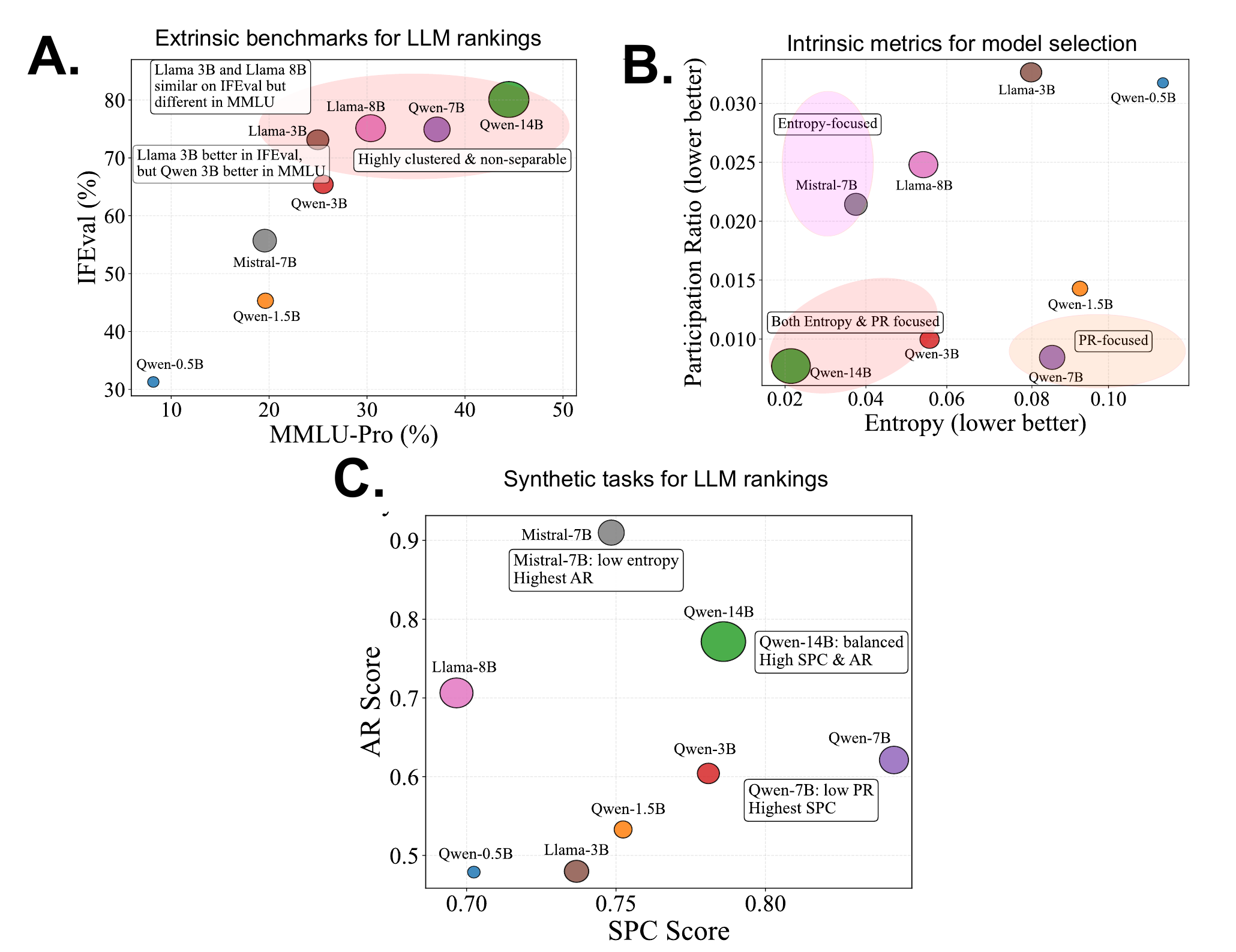}
    \caption{
    \textbf{Comparison of (A) extrinsic, (B) intrinsic, and (C) synthetic metrics for LLM ranking.} 
    \textbf{(A)} MMLU-Pro vs.\ IFEval shows inconsistent rankings across models (e.g., LLaMA-3B vs.\ Qwen-3B), with a dense top-right cluster indicating poor separability under extrinsic evaluation. 
    \textbf{(B)} Intrinsic metrics (entropy, participation ratio, effective rank) clearly separate models and offer actionable guidance for selecting models based on stability, activation diversity, or representational richness.
    \textbf{(C)} Synthetic tasks (SPC, AR) spread models along two largely independent axes that align with intrinsic properties.
    } \label{fig:motivation_result}
\end{figure}

\subsection*{Latent Performance Profiling reveals hidden differences}
To address these shortcomings, we introduce LPP as an intrinsic evaluation framework. Rather than focusing on what answer a model outputs, LPP examines how the model's certainty and representations evolve internally as it processes information. \color{black}We define a set of task-agnostic metrics (Table~\ref{tab:metrics}) computed from the model's hidden states and next-token probabilities that serve as proxies for core aspects of its internal computation. In this work, we focus on three such metrics: the \textit{uncertainty floor}, defined as the minimum, over prefix lengths, of the prompt-averaged next-token entropy (capturing the model's confidence floor); the \textit{effective rank} (ER), which estimates the active dimensionality of the hidden representations; and the \textit{participation ratio} (PR), which measures how evenly the variance of activations is distributed across dimensions. These metrics, aggregated per model (specifically using the extremal values: minimum entropy, maximum ER, and maximum PR), provide a compact ``latent profile'' that characterizes the model's confidence regime and representational breadth under generic inputs. We elaborate on these metrics in the \textit{Materials and Methods} section.

\input{tables/metrics}

\color{black}
Using LPP, we profile eight open-source models and find striking latent differences even when benchmark-based external performances are similar. For instance, Qwen-3B and Mistral-7B achieve comparable accuracy on the BBH benchmark (Figure~\ref{fig:results}A), yet they diverge markedly in their internal metrics (Figure~\ref{fig:results}B). Qwen-3B has a much lower participation ratio and effective rank than Mistral-7B, indicating that Qwen-3B's activations reside in a more compact, lower-dimensional subspace despite solving the same tasks. Mistral-7B, by contrast, uses a higher-dimensional and more diffuse representation. In other words, these two models appear equally capable based on external scores, but LPP reveals that they employ distinct internal strategies to reach those answers. Figure~\ref{fig:motivation_result} illustrates this phenomenon more broadly: models with similar benchmark scores can exhibit very different latent signatures. Traditional metrics alone cannot distinguish whether performance gains arise from fundamentally improved latent computation or merely from superficial changes, such as scale or fine-tuning.

\begin{figure}[t!]
    \centering
    \includegraphics[width=1.05\linewidth]{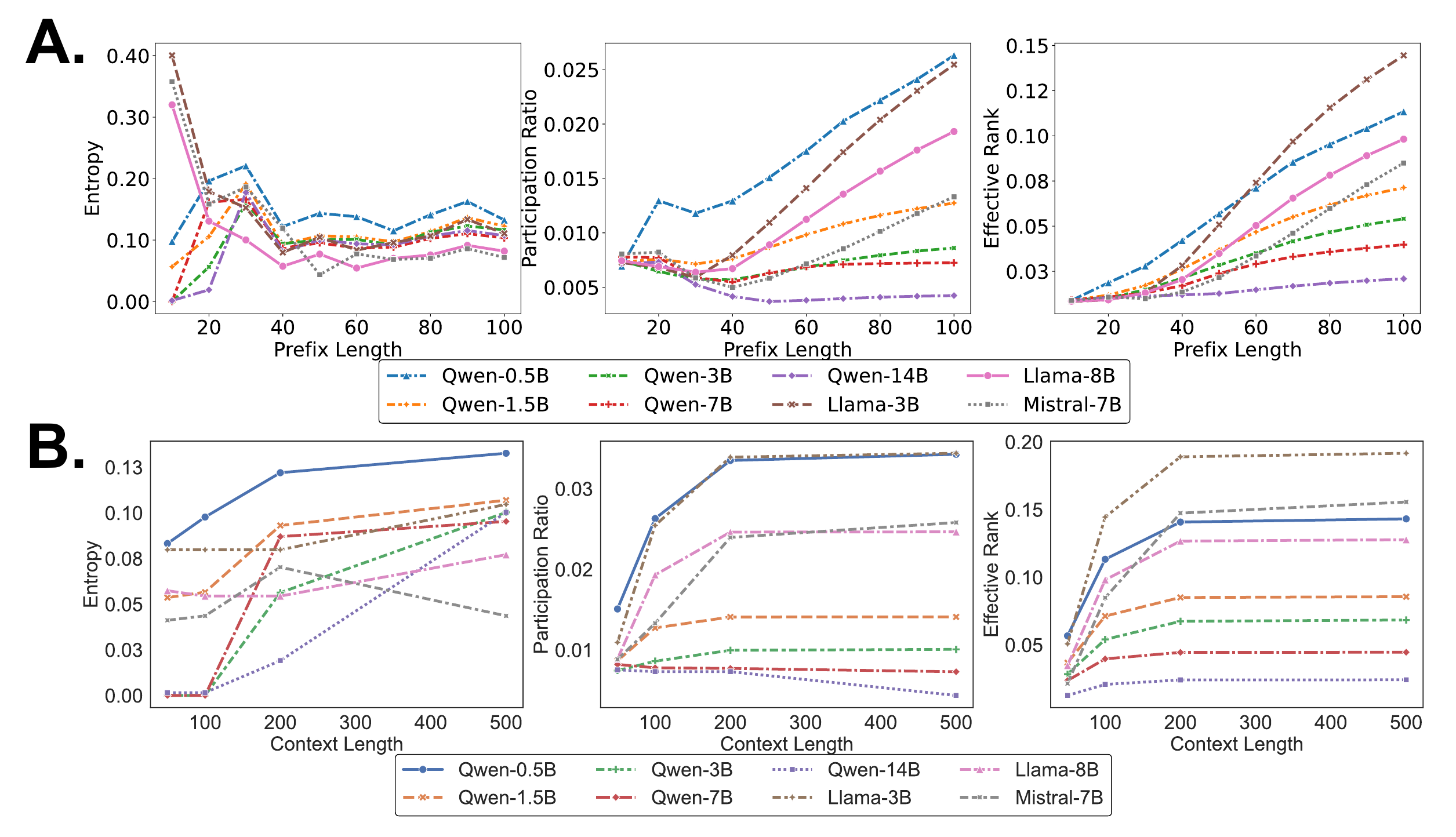}
    \caption{\textbf{Sensitivity of LPP metrics to (A) prefix length, and (B) context length.} PR and ER vary smoothly with prefix and context length, and the ordering of models is largely preserved for context lengths of 100 tokens and above. Entropy is more protocol-dependent: its trajectory and floor differ across models and context lengths. Values at a given context length are computed on that context's prefix grid and need not equal the profile values in Figure~\ref{fig:results}B.} \label{fig:sensitivity1}
\end{figure}

\color{black}
Examining LPP metrics across varying prefix and context lengths (Figure~\ref{fig:sensitivity1}) further shows that models differ in how their latent metrics evolve with context. \color{black} Across the eight models, the entropy floor decreases with model size (Pearson $r=-0.85$ with log parameter count), whereas PR and ER are less size-dependent ($r=-0.51$ and $-0.41$). In a two-dimensional entropy-PR plane, each model occupies a distinct region reflecting its calibration-compression trade-off. Qwen-14B and Qwen-3B, for example, combine low entropy with low PR, indicating highly confident predictions from very compact representations. Mistral-7B achieves the second lowest entropy among the evaluated models, indicating strong confidence, but maintains a moderate PR, whereas Qwen-7B attains the second lowest PR (highly compressed representations), but at the cost of higher entropy. In contrast, smaller models like LLaMA-3B and Qwen-0.5B exhibit both high entropy and high PR, corresponding to diffuse and uncertain latent states. Crucially, none of these differences is apparent from benchmark accuracy alone. For instance, LLaMA-3B and LLaMA-8B reach nearly identical IFEval scores, yet LLaMA-3B has a higher entropy floor and operates in a higher-dimensional latent space than its 8B counterpart. Together, these observations establish LPP as a principled tool for disentangling performance ambiguities and exposing latent characteristics that conventional evaluations overlook.

\subsection*{Intrinsic metrics versus extrinsic performance}

We next assess the extent to which traditional benchmark scores reflect the latent properties captured by LPP. Across the eight models, we compute Pearson correlations between their LPP metrics (entropy floor, max-ER, max-PR) and their MMLU-Pro, BBH, and IFEval scores (Figure~\ref{fig:results}D). The resulting correlations across three benchmarks and three intrinsic metrics range from roughly $-0.75$ to $-0.32$, with most correlations not statistically significant ($p$-value > 0.05). All nine correlations are negative, and seven lie between $-0.65$ and $-0.75$ ($p$ between 0.03 and 0.08), but with eight models only the entropy--BBH correlation reaches significance ($p=0.03$, unadjusted). Benchmark scores are therefore moderately but imprecisely associated with LPP metrics and do not determine a model's latent profile. For example, a high-performing model on BBH does not necessarily exhibit a low entropy floor or a compact latent space, and vice versa.

\color{black}

This partial association indicates that benchmark scores and LPP metrics carry related but distinct information, rather than that benchmarks are flawed. Extrinsic evaluations capture task-specific success but not attributes such as confidence management, internal diversity, or robustness to distribution shifts --- precisely the qualities that entropy, PR, and ER quantify. We therefore treat LPP as complementary to, not a replacement for, benchmark accuracy. As a result, purely outcome-based metrics cannot explain why certain models generalize better to unforeseen scenarios or why they might fail in settings prioritizing interpretability and robustness, despite strong validation performance. In our study, models with nearly identical accuracy often differed substantially in their latent profiles (Figures~\ref{fig:results}A vs. \ref{fig:results}B), underscoring the gap between the extrinsic and intrinsic measures. This finding motivates the development of new evaluations that specifically target the latent traits highlighted by LPP.

\color{black}

\subsection*{LPP-informed tasks for model evaluation}

To probe the latent competencies revealed by LPP, we design two diagnostic tasks directly inspired by the intrinsic dynamics of entropy, PR, and ER. The first task, \textit{Ambiguous Reasoning} (AR), presents the model with a prefix containing a lexical ambiguity followed by a disambiguating cue, challenging it to recognize the ambiguity and select the correct resolution. This task specifically tests the model's ability to recognize uncertainty and calibrate entropy when resolving ambiguity. The second task, \textit{Symbolic Pattern Completion} (SPC), presents the model with a synthetic sequence governed by a hidden rule (for example, an alternating or mirrored pattern) and requires it to predict the subsequent symbols accurately. This evaluates the model's capacity for structural compression and pattern generalization, which depend on its representational efficiency (PR and ER). Representative examples of both task types are provided in Table~\ref{tab:examples}. Each task comprises 100 prompts built from templates rather than existing corpora, to reduce the influence of memorized content. The AR set contains 20 distinct ambiguity items, each presented five times; chance accuracy is approximately 0.5.

\input{tables/tasks}

Across models, performance on these LPP-informed tasks aligns closely with the models' latent profiles. In the AR task, models with lower entropy floors succeed more often at resolving ambiguities. Accordingly, AR accuracy is strongly \textit{negatively} correlated with minimum entropy across the model pool ($r=-0.79$, $p=0.02$, unadjusted, $n=8$; Figure~\ref{fig:results}E); models that maintain low entropy in the face of uncertainty tend to resolve ambiguous prompts correctly, whereas models that exhibit higher irreducible entropy struggle with ambiguity. \color{black}The strength of this association depends on the prefix schedule: on the 100-, 200- and 500-token sensitivity grids (Figure~\ref{fig:sensitivity1}B), it is $r=-0.44$, $-0.60$ and $-0.87$, respectively. \color{black} In the SPC task, we observe a similarly tight link to representational metrics. Models with more compact and less redundant latent representations achieve higher SPC scores, yielding negative correlations between SPC performance and both PR ($r=-0.79$, $p=0.02$) and ER ($r=-0.71$, $p=0.047$), unadjusted for multiple comparisons. For example, Qwen-7B, which has one of the most compact latent spaces (low PR and ER), achieves the highest SPC F1 among the tested models. Similarly, Mistral-7B's strong showing on AR occurs even though it is not the top performer on any traditional benchmark, highlighting how LPP-based evaluation can surface capabilities that leaderboards miss.

These intrinsic tasks also shed light on model scaling from a process perspective. AR performance generally improves with model size within the evaluated families, whereas SPC performance is non-monotonic. Qwen-7B achieves the highest SPC score. Smaller models in our study tend to be under-confident (higher entropy) and rely on more diffuse activations (higher PR/ER). In contrast, the larger models combine low uncertainty with structured, lower-dimensional representations, enabling strong performance. \textcolor{black}{To probe whether these results extend beyond these model sizes, we additionally evaluated Qwen2.5-32B (4-bit), which scores 0.76 on AR and 0.844 on SPC. It matches Qwen-14B on AR and is comparable to Qwen-7B on SPC (0.844 vs. 0.842), consistent with the non-monotonic SPC trend observed in the main pool\footnote{Unless explicitly stated otherwise, model rankings refer to the eight-model main analysis and exclude this exploratory 32B evaluation.}}. Interestingly, AR and SPC performance are largely uncorrelated across models, indicating that each task probes a distinct dimension of latent reasoning (calibration vs. compression). Consistent with this, we find that AR results correlate tightly with entropy-based metrics, whereas SPC results correlate with rank-based metrics (PR, ER). The cross-correlation between AR-related and SPC-related features is low and not statistically significant. This confirms that the two tasks provide complementary evaluations of model internals.

Crucially, the LPP-guided tasks separate model capabilities more clearly than standard benchmarks. In a two-dimensional plot of AR versus SPC performance (Figure~\ref{fig:motivation_result}C), models occupy distinct regions corresponding to their latent profiles. Mistral-7B, with its low entropy floor, dominates the AR axis (excelling at ambiguity resolution), whereas Qwen-7B, with an extremely low PR, leads on the SPC axis (excelling at pattern completion). Qwen-14B balances both traits, achieving strong performance on both tasks. This separation is visibly clearer than on conventional benchmarks (Figure~\ref{fig:motivation_result}A versus C). In fact, models that appear tied in extrinsic accuracy are well-distinguished by their AR and SPC scores, suggesting that these tasks capture intrinsic reasoning and representational traits rather than superficial effects of scale. Moreover, because AR and SPC target different latent dimensions, a model's strengths in one do not guarantee strengths in the other, reinforcing the value of a multifaceted evaluation.

Taken together, our results demonstrate the practical utility of latent profiling. While extrinsic metrics alone cannot predict a model's reliability on nuanced tasks, LPP exposes latent traits that help explain and even anticipate model behavior. By guiding the design of targeted evaluations (like AR and SPC), LPP offers a path toward more robust benchmarks that probe the mechanisms \textit{underlying} performance rather than merely the end results.

\color{black}
Because AR and SPC are our own constructions, we further ask whether LPP metrics predict behavior on tasks and judgments we did not design. Entropy is negatively correlated with accuracy on TruthfulQA-MC~\citep{lin2021truthfulqa}, an independent hallucination benchmark (Spearman $\rho=-0.80$, $p=0.017$, $n=8$), although this association weakens once model size is controlled for (partial $\rho=-0.29$, $p=0.48$). Effective rank shows a negative but non-significant association with performance on Dyck-languages, a symbolic sequence-completion task from BIG-Bench Hard~\citep{suzgun2023challenging} that mirrors SPC ($\rho=-0.69$, $p=0.058$). Using a strong external model (Claude Opus 4.8) as an independent judge~\citep{zheng2023judging,gilardi2023chatgpt}, lower-entropy models are rated as more reliable ($\rho=-0.73$, $p=0.04$), and the judge's model-level AR scores rank the eight models similarly to our automatic scorer ($\rho=0.93$), although the judge's absolute scores are lower by 0.05--0.29. We report these external validations and the outlier-dimension robustness check in \textit{Supplementary} Sections 2.5 and 2.6.

\color{black}

\section*{Discussion}

\subsection*{LPP complements traditional benchmarks}
Leaderboards provide a convenient snapshot of performance, but our findings show that they can obscure critical facets of model behavior. In our evaluations, models with nearly identical scores on MMLU-Pro, IFEval, or BBH often exhibited very different internal profiles (Figure~\ref{fig:results}), indicating that high test accuracy can mask underlying vulnerabilities. Indeed, benchmark scores are susceptible to inflation due to data overlap, test-set memorization, or benchmark-specific fine-tuning, making it difficult to distinguish genuine reasoning ability from exposure-driven artifacts. This issue is exacerbated at scale: large models that differ substantially in calibration, robustness, or representational geometry can nonetheless achieve deceptively similar leaderboard scores. Aggregate metrics collapse diverse behaviors into a single number, hiding failure modes such as miscalibrated uncertainty, brittle intermediate representations, or over-expanded activations that are sensitive to context shifts. Consequently, a high ranking on a public benchmark often provides an incomplete and sometimes distorted view of a model's true capabilities. LPP directly addresses these blind spots by quantifying internal properties that conventional benchmarks cannot observe. The entropy, PR, and ER metrics are computed on unlabeled text and do not depend on any benchmark's test set. By measuring a model's uncertainty floor, degree of latent compression, and active dimensionality, LPP reveals whether the model's internal mechanisms are conducive to robust generalization, even when its outward accuracy appears strong. 

In our study, we observed that models of comparable size from different families can have divergent intrinsic profiles (for example, Qwen-7B and Mistral-7B), and that profiles also vary substantially with scale within a family. These nuances are invisible to traditional benchmarks but become clear through LPP analysis. Furthermore, LPP offers an explanation for why extrinsic metrics and latent behavior can diverge: models with similar benchmark scores may rely on entirely different representational strategies, such as compact versus diffuse manifolds, which are not distinguished by accuracy alone. Notably, we also find that model rankings based on LPP metrics are largely consistent when means or medians replace the extremal values (Spearman rank agreement 0.81--0.98; \textit{Supplementary} Figure~5), although aggregating ER by its minimum reverses the ordering; in addition, different aggregation strategies can highlight distinct aspects of internal behavior. For instance, minimum entropy emphasizes best-case calibration, whereas average PR reflects the general representational spread. This suggests that a more informative way to summarize LPP results is to use a vectorized profile of multiple statistical summaries per metric, yielding a richer and more interpretable latent signature. By complementing extrinsic scores with such intrinsic diagnostics, LPP adds a principled, process-aware layer to model evaluation.

\subsection*{Relation to matrix entropy and interpretability methods}

\color{black}

LPP draws conceptual inspiration from earlier work on matrix entropy~\citep{wei2024large} and rank-based evaluations of neural activations~\citep{ansuini2019intrinsic,jha2025spectral,skean2025layer}. However, unlike classical matrix entropy measures that are often limited to single snapshots of layer activations or require heuristic normalization, LPP incorporates multiple prefix lengths to construct a stable latent signature. Moreover, while entropy alone captures the uncertainty in a model's predictions, it does not reveal how those predictions are supported by internal representations. PR and ER extend beyond this by quantifying the structure and dimensionality of latent spaces, offering a richer picture of how information is stored and transformed. Compared to mechanistic interpretability techniques~\citep{bereska2024mechanistic,dreyer2025mechanistic}, which aim to dissect individual neurons, attention heads, or circuits, LPP provides a more scalable and model-agnostic diagnostic. Circuit-level approaches~\citep{olah2020zoom,elhage2021mathematical,wang2023interpretability,conmy2023towards} can yield powerful insights but often require manual tracing or extensive intervention (e.g., patching or ablation), making them hard to generalize across architectures or tasks. In contrast, LPP requires no supervision or access to model internals beyond activations, enabling systematic comparison across diverse models without task-specific adaptation. It provides coarse-grained but actionable insight into global representational behavior, enabling us to triage which models or layers warrant further mechanistic inspection. Importantly, PR and ER offer distinct advantages over probing-based evaluations. Probing typically focuses on whether particular information (e.g., syntax, world knowledge) is linearly decodable from intermediate representations, which can be misleading: a model may store information in a way that is useful for internal computation but not for external probing. PR and ER sidestep this limitation by measuring the intrinsic geometry of latent spaces, independent of downstream tasks or classifiers. They capture the compressibility and redundancy of activations, properties that directly impact a model's generalization, efficiency, and susceptibility to adversarial inputs. Thus, LPP augments and extends the existing toolbox of interpretability and diagnostic methods by offering complementary, representation-level insights that are robust, lightweight, and general-purpose. We situate LPP more fully within the literature on uncertainty calibration, representational geometry, and intrinsic dimensionality, and circuit-level interpretability in \textit{Supplementary} Section 2.8.

\color{black}

\subsection*{Stable latent patterns and their origins}
Our analyses also shed light on the source of the observed LPP patterns. The entropy, PR, and ER metrics we measured appear to arise from fundamental architectural and representational properties, rather than from idiosyncratic performance differences. In particular, across the evaluated context lengths (Figure~\ref{fig:sensitivity1}B), entropy exhibits model-specific trajectories, with some curves approaching a plateau as context increases. 
\color{black} The entropy floor also shifts with the prefix schedule for several models, so it is best read as a property of a model under a fixed measurement protocol. \color{black}
The representational metrics, such as PR and ER, exhibit similarly structured behavior. As shown in Figure~\ref{fig:sensitivity1}B, ER generally increases with context length, whereas PR trends are model-dependent; for instance, Qwen-14B's PR decreases slightly at longer contexts. This indicates that as an input sequence grows, models systematically expand the dimensionality of their latent representations, presumably to encode the increasing complexity or information in longer contexts. Notably, this pattern holds across different model families (LLaMA, Qwen, Mistral), hinting at a shared underlying principle of decoder-only transformers: their hidden representations evolve from more compressed manifolds over short prefixes to richer, higher-dimensional subspaces as more context accumulates. At the same time, different architectures allocate capacity in different ways. For example, Qwen-7B maintains an extremely low PR even at the maximum context length (indicating a very tight, sparse representation), whereas LLaMA-3B's PR grows much larger, indicating a more diffuse activation pattern. In effect, Qwen-7B aggressively compresses its latent space (perhaps reflecting a more modular or efficient encoding strategy), while LLaMA-3B rapidly expands its latent space, which could make it more expressive but potentially more redundant. Such differences likely stem from architectural choices or training regimes that favor certain encoding strategies over others.

Examining the LPP metrics across model depths further reveals a shared mid-layer bottleneck. In all eight models, PR and ER fall to their lower bound in the middle layers and rise again in the final layers (see \textit{Supplementary} Figure~4). The Qwen models additionally show high PR and ER in the early layers, giving an ``hourglass'' profile, whereas LLaMA-3B, LLaMA-8B, and Mistral-7B start near the lower bound and rise only at the end. The mid-layer compression and late re-expansion, therefore, appear to be common to the decoder-only models we studied, while the early-layer expansion is family-specific. The consistency of these profiles, along with the context-length trends discussed above, indicates that LPP metrics capture fundamental aspects of model design, such as attention allocation, activation distribution, and latent bottlenecks. Model rankings are stable across sample sizes and, for ER and PR, across calibration datasets (\textit{Supplementary} Figures~1--3); entropy rankings are more dataset-dependent when very short prefixes are included. This supports the view that the representational metrics reflect model properties rather than artifacts of a particular dataset. In summary, the latent profiles revealed by LPP appear to stem from the models' core architecture and training, providing insight into how modern LLMs manage information flow and complexity internally.

\subsection*{Implications for model selection and monitoring}
Beyond analysis, LPP offers practical guidance for model selection and deployment. By characterizing models in terms of their latent profiles, we can identify which models are inherently better suited for certain applications. For example, Figure~\ref{fig:motivation_result}B highlights several latent ``archetypes.'' Models with low entropy floors (such as Qwen-14B or Mistral-7B) produce narrower, more confident predictive distributions and performed best on our ambiguity-resolution task. Confidence is not calibration, however: modern networks can be confidently wrong, particularly under distribution shift~\citep{guo2017calibration,ovadia2019can}, and users tend to over-trust confident LLM answers~\citep{steyvers2025large}, so in safety-critical settings the entropy floor should be read alongside accuracy-based calibration measures. In contrast, models optimized for low PR (such as Qwen-7B, with extremely compact representations) excel at tasks involving algorithmic pattern recognition or symbolic manipulation, where a highly compressed internal representation is beneficial. Meanwhile, a model like Qwen-14B, which balances low entropy and low PR, offers more general-purpose stability, performing reliably across a broad range of tasks. These distinctions are not readily apparent from standard metrics alone, but LPP makes them explicit. Therefore, when choosing an LLM for a particular use case, one can consider its LPP profile (Figure~\ref{fig:results}B) in addition to its task accuracy: \color{black} for instance, preferring a model with a low entropy floor for tasks that require resolving ambiguous inputs, as in our AR task, or a model with a low participation ratio for symbolic pattern completion. \color{black}

In deployment, LPP metrics can also serve as valuable monitors for model behavior. Because entropy, PR, and ER respond in predictable ways to increasing context (Figure~\ref{fig:sensitivity1}B) and other input perturbations (Figure~\ref{fig:sensitivity1}A and \textit{Supplementary} Figures~1--3), deviations from these expected patterns can act as early-warning signals. For example, if a deployed model that normally maintains a certain entropy floor suddenly exhibits a rising trend in entropy on incoming data, it may indicate domain drift or novel inputs that undermine the model's calibration. Similarly, an unexpected spike in effective rank or participation ratio may indicate that the model's internal representations are straying into an anomalously diffuse regime, perhaps due to adversarial inputs or to the model entering a corner of its parameter space that is not well covered during training. Such latent shifts could presage performance degradation or unsafe behavior long before these issues surface in output accuracy. By continuously tracking LPP metrics during inference, practitioners could detect these warning signs and intervene (e.g., by model retraining or input filtering) before failures occur. In this way, LPP enables interpretable model telemetry, providing real-time insight into the model's ``health indicators'' and thereby supporting more stable and accountable deployment. While this work focuses on three core metrics --- entropy, participation ratio, and effective rank --- the LPP framework is extensible, and future work may introduce additional metrics that target other facets of latent computation, such as attention dispersion, context sensitivity, or activation sparsity. Overall, incorporating LPP-based criteria into model selection and monitoring protocols can help ensure that we not only pick models that perform well on paper but also those that exhibit desirable internal behavior during operation.

\section*{Materials and Methods}
\label{sec:method}

In this section, we formally define the LPP metrics, describe how they are computed, and outline the procedures used for constructing and evaluating synthetic tasks.

\subsection*{Formal definition of LPP metrics}

\color{black}

We consider three latent metrics that capture distinct aspects of a model's internal processing (see Table~\ref{tab:metrics}). Unless stated otherwise, entropy is computed from the next-token distribution, and representational metrics are computed from last-layer hidden states, using 100 task-agnostic Alpaca~\citep{alpaca} prompts. For a hidden activation tensor $h=h_{\ell}(x;\theta)$ at layer $\ell$ and token sequence $x$, we mean-center $h$ across tokens and let $C=\mathrm{Cov}(h)$ denote the empirical covariance matrix of the token representations pooled across all calibration prompts. Let $\{\lambda_i\}_{i=1}^{K}$ be the eigenvalues of $C$, where $K=\min(N,d)$ for $N$ pooled token representations and hidden dimension $d$, and define normalized eigenvalues $\tilde{\lambda}_i=\lambda_i/\sum_{j}\lambda_j$. All three metrics are normalized to the unit interval so that values are comparable across models: entropy by $\log|\mathcal{V}|$, where $\mathcal{V}$ is the vocabulary, and ER and PR by $K$. Because transformer activations can contain a few high-variance ``outlier'' dimensions~\citep{puccetti-etal-2022-outlier,dettmers2022gptint}, we verify that our rankings and conclusions are unchanged when these dimensions are robustly down-weighted (\textit{Supplementary} Section 2.6).

\color{black}

\noindent\fcolorbox{black}{blue!10}{%
\minipage[t]{\dimexpr\linewidth-2\fboxsep-2\fboxrule\relax}
\textbf{Uncertainty floor (Entropy).}
For token position $t$ and prefix $x_{\le t}$, we define the normalized next-token entropy as
\[
H_t=-\frac{1}{\log|\mathcal{V}|}\sum_{v\in\mathcal{V}} P_\theta(v\mid x_{\le t}) \log P_\theta(v\mid x_{\le t}), \qquad H_t\in[0,1].
\]
We average $H_t$ over the calibration prompts for each prefix length $t$ and report the minimum of these averages across prefix lengths (the entropy floor). A higher-entropy floor indicates persistent uncertainty even as contextual cues accumulate, whereas a lower but well-calibrated floor corresponds to models that become confident only when evidence is sufficient. Extremely low floors may signal premature overconfidence.
\endminipage}

\vspace{0.6em}

\noindent\fcolorbox{black}{green!10}{%
\minipage[t]{\dimexpr\linewidth-2\fboxsep-2\fboxrule\relax}
\textbf{Representational expansiveness (Effective Rank).}
With covariance $C$ and normalized eigenvalues $\{\tilde{\lambda}_i\}_{i=1}^{K}$, we compute the normalized effective rank as
\[
\mathrm{ER}=\frac{1}{K}\exp\!\left(-\sum_{i=1}^{K} \tilde{\lambda}_i \log \tilde{\lambda}_i\right), \qquad \mathrm{ER}\in\left[\tfrac{1}{K},1\right].
\]
ER estimates the fraction of active dimensions in the hidden representation. Excessively high ER correlates with diffuse, over-expanded representations and degraded task performance, whereas moderate ER reflects rich but structured activity.
\endminipage}

\vspace{0.6em}

\noindent\fcolorbox{black}{blue!10}{%
\minipage[t]{\dimexpr\linewidth-2\fboxsep-2\fboxrule\relax}
\textbf{Variance distribution (Participation Ratio).}
Given eigenvalues $\{\lambda_i\}_{i=1}^{K}$ of $C$, we define the normalized participation ratio
\[
\mathrm{PR}=\frac{1}{K}\,\frac{\left( \sum_{i=1}^{K} \lambda_i \right)^2}{\sum_{i=1}^{K} \lambda_i^2}, \qquad \mathrm{PR}\in\left[\tfrac{1}{K},1\right].
\]
PR quantifies how evenly variance is distributed across dimensions. A very high PR suggests diffuse activation and redundant capacity allocation, while a lower PR reflects more selective and compressed representations.
\endminipage}

\vspace{1em}

To summarize behavior across prefix lengths, we extract extrema that characterize failure-prone or stability-critical regimes:  
(i) minimum next-token entropy (entropy floor),  
(ii) maximum effective rank (max-ER), and  
(iii) maximum participation ratio (max-PR).  

These extrema capture the model's most confident states and its most expanded or most distributed representational modes. In practice, we compute LPP metrics over 100 Alpaca prompts, estimate from the last layer of each LLM, and aggregate them using the extrema above to obtain a compact latent profile for each model. We further elaborate on the sensitivity of LPP metrics to different aggregation functions in \textit{Supplementary} Figure~5.

\subsection*{Implementation details}

\color{black}

\textbf{Models and environment:} We evaluate eight publicly available instruction-tuned models from Hugging Face using the \texttt{transformers} library: Qwen2.5 at 0.5B, 1.5B, 3B, 7B, and 14B~\citep{qwen2025qwen2}, LLaMA-3.2-3B and LLaMA-3-8B~\citep{dubey2024llama}, and Mistral-7B-v0.2~\citep{jiang2023mistral7b}. External benchmark scores on MMLU-Pro, BBH, and IFEval are obtained from the Hugging Face leaderboard~\citep{open-llm-leaderboard-v2}; these are leaderboard-normalized scores, not raw accuracies. For AR and SPC generation, we use left padding and assign the EOS token as the padding token when it is absent. LPP metrics are computed with random seed 42 and batch size 1 on the first 100 Alpaca examples containing at least 128 tokens; entropy is computed from the raw next-token logits of a single forward pass (no decoding). AR and SPC are evaluated using greedy decoding, a batch size of 4, and seed 1337.

\noindent \textbf{Datasets and prompts for intrinsic metrics:}  Intrinsic metrics are computed on 100 task-agnostic texts from the Alpaca dataset; entropy and hidden-state metrics are obtained from the same forward passes. \color{black} For the LPP profile (Figure~\ref{fig:results}B), we evaluate prefixes of 20 to 150 tokens in steps of 10, using the same schedule for all eight models, and take the minimum entropy and the maximum PR and ER over these prefixes. We exclude the 10-token prefix, at which several Qwen models show near-zero entropy. In the sensitivity analysis (Figure~\ref{fig:sensitivity1}), a context length $C$ is evaluated at prefixes $C/10, 2C/10, \dots, C$. \color{black} Sensitivity analyses vary context length $\{50,100,200,500\}$, prefix length $\{10,20,30,40,50,60,70,80,90,100\}$, and sample size $\{10,100,500,1000\}$ across WikiText~\citep{merity2016pointer}, Dolly~\citep{DatabricksBlog2023DollyV2}, and Alpaca~\citep{alpaca}. All datasets are English-language corpora for autoregressive generation. \color{black} Although Alpaca is instruction-style text, relative model rankings for ER and PR are stable across these corpora, whereas entropy rankings are more dataset-dependent when very short prefixes are included (\textit{Supplementary} Section 2.1). \color{black}


\color{black}

\subsection*{Synthetic tasks motivated by LPP metrics}

To probe the latent behaviors revealed by LPP, we construct two synthetic evaluation tasks --- Ambiguous Reasoning and Symbolic Pattern Completion --- that isolate uncertainty dynamics and representational compression. These tasks are derived directly from the trends observed in entropy, PR, and ER, and avoid contamination by relying on procedurally generated templates. Algorithms 1 and 2 in \textit{Supplementary} Section 1 formalize the synthetic task generation process. Examples are provided in Table~\ref{tab:examples}. Each dataset contains 100 samples. \textbf{Ambiguous Reasoning (AR):}  
AR tests whether a model can (i) detect ambiguity in a prefix, and (ii) resolve it after receiving a disambiguating hint. \color{black} This design is motivated by the uncertainty-floor behavior captured by entropy: the prefix alone leaves the sense open, and the hint supplies the evidence needed to resolve it. We score only the final answer choice. \color{black} Prefixes, hints, and answer options are generated from templates; the set contains 20 distinct items, each presented five times. \textbf{Symbolic Pattern Completion (SPC):}  
SPC evaluates a model's ability to infer and extend symbolic rules in short sequences (e.g., alternation, mirroring, modular increments). These tasks reflect compression-based behavior associated with PR and ER: \color{black} strong models exhibit low PR and low ER (compact, selective representations). \color{black} Sequences are generated procedurally to avoid domain-specific priors and to focus on structural generalization. Both AR and SPC allow precise control over difficulty, sequence structure, and ambiguity type, enabling targeted assessment of the latent dimensions uncovered by LPP.

\subsection*{Evaluating LLMs on the AR and SPC tasks}

We evaluate both tasks using few-shot prompting with batched generation. Inputs are left-padded; when no pad token exists for a tokenizer, EOS is substituted to prevent misaligned logits. For AR, we use \texttt{max\_new\_tokens=16}; for SPC, \texttt{max\_new\_tokens=8}. Each prompt includes 10 demonstrations sampled once (seed 1337) from the task file itself; for AR, these are answer-format demonstrations only. \textbf{Metrics:} For SPC, we compute character-level F1, defined as the overlap of character counts between prediction and target (order-insensitive), averaged across samples; exact-match accuracy is provided in the repository. For AR, we compute answer-choice accuracy (chance $\approx 0.5$). All results are reported as dataset-level averages. Further analyses on the impact of the number of in-context examples on the model performance are highlighted in \textit{Supplementary} Figure~6.

\noindent All experiments are conducted on a single NVIDIA A100 GPU. Calculating LPP metrics on 100 samples takes 2--4 minutes, depending on model size. 


\section*{Limitations and future scope}
The present study is designed as a first, controlled demonstration of latent profiling, and its findings are best interpreted within that scope. Our analysis focuses on eight open-source, decoder-only models from three families (0.5B--14B parameters); with eight models and unadjusted tests, the observed associations are therefore indicative, and several LPP metrics, entropy in particular, co-vary with model size. Extending the analysis to larger and architecturally different systems is a natural next step. Because LPP operates on hidden states and output logits, it currently applies to models whose internals are accessible, making it readily usable for the growing open-source ecosystem. Latent profiles reflect the measurement protocol (calibration corpus, context length, layer, and aggregation function): \color{black} PR and ER rankings are broadly stable across calibration corpora and context lengths of 100 tokens or more, whereas the entropy floor is more protocol-dependent; absolute values are most meaningful when compared under a fixed protocol. \color{black} The AR and SPC tasks are intentionally compact diagnostic probes, and our external validation on two independent benchmarks and an LLM judge offers encouraging initial support, with broader benchmarks and human evaluation likely to strengthen it further. Finally, the relationships we report are associational; establishing whether entropy, PR, and ER causally shape downstream behavior, for example, through targeted interventions, is a promising direction for future work.

\section*{Data Availability}

This study analyzes publicly available text corpora, publicly released model weights, and published benchmark scores, and generates two synthetic datasets. No human-subject or personally identifiable data were used or generated. All data generated in this study are stored in the \texttt{data/} directory of the public GitHub repository \url{https://github.com/parmanu-lcs2/lpp} and are freely available without access restrictions under the MIT license of the repository.

\noindent \textbf{Generated data.} The Ambiguous Reasoning (AR) and Symbolic Pattern Completion (SPC) datasets each contain 100 samples. They are stored in JSON Lines format with the input prompt, the gold target, and item metadata (for AR: prefix, candidate senses, hint, and correct option; for SPC: generating rule and parameters). The few-shot demonstrations used in evaluation are drawn from these files with a fixed random seed (see Code Availability).

\noindent The repository also contains the source data underlying Figures~2--4 and \textit{Supplementary} Figures~1--9. A table in the repository's README maps each file to the figure it supports.

\noindent \textbf{Third-party data.} Intrinsic metrics were computed on three public corpora: Stanford Alpaca~\citep{alpaca} (\url{https://huggingface.co/datasets/tatsu-lab/alpaca}), Databricks Dolly-15k~\citep{DatabricksBlog2023DollyV2} (\url{https://huggingface.co/datasets/databricks/databricks-dolly-15k}) and WikiText-103~\citep{merity2016pointer} (\url{https://huggingface.co/datasets/Salesforce/wikitext}). The prompt set for each corpus is selected deterministically by the released code, as the first examples that contain at least 128 tokens, so it can be regenerated exactly. External validation used TruthfulQA~\citep{lin2021truthfulqa} (\url{https://huggingface.co/datasets/truthfulqa/truthful_qa}) and the Dyck-languages task of BIG-Bench Hard~\citep{suzgun2023challenging} (\url{https://github.com/suzgunmirac/BIG-Bench-Hard}). Extrinsic scores on MMLU-Pro, BBH, and IFEval were taken from the Open LLM Leaderboard v2~\citep{open-llm-leaderboard-v2} through its public results dataset (\url{https://huggingface.co/datasets/open-llm-leaderboard/contents}; accessed October 2025). All evaluated models are publicly available on the Hugging Face Hub, and their identifiers are listed in the repository. These third-party resources remain subject to their original licenses.

\noindent \textbf{Requests.} Any additional data supporting the findings of this study are available from the corresponding author, Tanmoy Chakraborty (\href{mailto:tanchak@iitd.ac.in}{tanchak@iitd.ac.in}), upon reasonable request.

\section*{Code Availability}

All custom code used in this study is publicly available without restriction at \url{https://github.com/parmanu-lcs2/lpp} under the MIT license. The version used for this paper corresponds to release v1.1. The repository contains:
\begin{itemize}
    \item \texttt{lpp.py}, which computes the LPP metrics for any Hugging Face causal language model;
    \item \texttt{aggregate.py}, which aggregates per-prefix metrics into model-level profiles;
    \item \texttt{eval\_ar\_fewshot.py} and \texttt{eval\_spc\_fewshot.py}, together with the driver script \texttt{run\_ar\_spc.sh}, for the synthetic-task evaluations;
    \item Jupyter notebooks covering the result compilation, ablation studies, statistical tests, and generating visualizations.
\end{itemize}

\noindent \textbf{Software.} The code is written in Python ($\geq$3.10) and uses PyTorch, Hugging Face Transformers, Datasets, Accelerate, bitsandbytes (for 4-bit loading), NumPy, SciPy, pandas, scikit-learn, Matplotlib, and Seaborn; the minimum versions are listed in \texttt{requirements.txt}. The independent judge was Claude Opus 4.8 (API model identifier \texttt{claude-opus-4-8}), accessed through the Anthropic Python SDK. Its prompts are given in \textit{Supplementary} Box~1 and in the GitHub repository. All experiments were run on a single NVIDIA A100 GPU.

\section*{Author Contributions}
T.C. conceived the idea. T.C., A.S., S.B., P.P.C., A.C., S.C., P.P.D., L.D., R.S., and M.V. helped in refining the idea. A.S. ran all the experiments. T.C. and A.S. analyzed the results and prepared the initial manuscript. R.S. and M.V. reviewed the analyses and statistical reporting and contributed to revising the interpretation of the results. T.C., A.S., S.B., P.P.C., A.C., S.C., P.P.D., L.D., R.S., and M.V. contributed to finalizing the manuscript. 


\section*{Competing Interests}
The authors have no competing interests. 

\section*{Acknowledgments}

T.C. acknowledges the support of the Rajiv Khemani Young Faculty Chair Professorship
in Artificial Intelligence and the NVIDIA Academic Grant Program. R.S. and M.V. acknowledge support from the Srijan Centre of Excellence on Generative AI at IIT Jodhpur, funded by the IndiaAI Mission (MeitY) and Meta. 

\thispagestyle{empty}

\nocite{*}

\bibliography{main_comb}

\clearpage

\setcounter{section}{0}
\setcounter{subsection}{0}
\setcounter{subsubsection}{0}
\setcounter{figure}{0}
\setcounter{table}{0}
\setcounter{algorithm}{0}

\begin{center}
{\color{color1}\Large\sffamily\bfseries Supplementary Material}
\end{center}

\renewcommand{\figurename}{Supplementary Figure}
\renewcommand{\tablename}{Supplementary Table}

\section{Materials and Methods}

\subsection{Generation of synthetic tasks}

To evaluate how intrinsic latent properties of LLMs relate to their reasoning capabilities, we design two synthetic diagnostic tasks: Ambiguous Reasoning (AR) and Symbolic Pattern Completion (SPC). These tasks are explicitly constructed to probe the predictive confidence and representational compactness of language models, guided by LPP metrics such as entropy, effective rank (ER), and participation ratio (PR).

\subsubsection{Ambiguous Reasoning (AR)}
The AR task is designed to assess a model's ability to resolve lexical ambiguity. It requires the model to choose between two plausible interpretations of a short prefix, given a disambiguating hint. Items are generated from templates as follows:

\begin{itemize}
    \item We define a small set of lexical ambiguity types, each built around a word with two common senses (for example, \textit{date} as a calendar day or a fruit, or \textit{bank} as a financial institution or a river bank). The released set covers five types of ambiguity.
    \item For each type, a short prefix (three to five words) is written so that it ends just before the ambiguous word and is compatible with both senses.
    \item Two candidate continuations are provided, one per sense (options A and B).
    \item A one-sentence hint is added that supports exactly one of the two senses; the position of the correct option is balanced (48\% A, 52\% B).
    \item The model is asked to label the prefix as ambiguous or unambiguous and to choose the correct option (A or B) based on the hint. Only the answer choice is scored.
\end{itemize}

The released AR set contains 100 prompts, each drawn from 20 distinct items and presented 5 times, so the chance accuracy is approximately 0.5. The task is motivated by the entropy floor: a model that remains uncertain between the two senses until the hint arrives, and then commits to the supported sense, should select the correct option.

\begin{algorithm}[H]
\caption{Generate AR Tasks}
\begin{algorithmic}[1]
\State \textbf{Input:} Ambiguity types $\mathcal{A}$ (word with two senses), prefix templates, hint templates $\mathcal{H}$, repetitions $r$
\State \textbf{Output:} Ambiguous Reasoning task set $\mathcal{T}_{AR}$

\State Initialize task set $\mathcal{T}_{AR} \gets \emptyset$
\ForAll{ambiguity type $a$ in $\mathcal{A}$}
    \ForAll{prefix template $p$ of type $a$}
        \State Form the two sense-specific continuations $s_1$, $s_2$
        \State Select the target sense and generate a hint $h \sim \mathcal{H}$ that supports it
        \State Assign $s_1$, $s_2$ to options A and B
        \State Format input prompt:
        \Statex \quad \texttt{Prefix:} $p$, \texttt{Options:} A. $s_1$, B. $s_2$, \texttt{Hint:} $h$
        \State Append to $\mathcal{T}_{AR}$, $r$ times, with the correct option as the answer
    \EndFor
\EndFor
\State \Return $\mathcal{T}_{AR}$
\end{algorithmic}
\end{algorithm}

\subsubsection{Symbolic Pattern Completion (SPC)}
The SPC task probes whether models can extract, compress, and generalize symbolic rules, emulating a low-dimensional representation strategy. It is constructed as follows:

\begin{itemize}
    \item A library of five symbolic rules is defined over letters, digits, and brackets: periodic repetition (e.g., AZLAZL), alternation (e.g., DYDY), mirroring (e.g., TOOTTOOT), bracket patterns (e.g., ()()()\{\}), and modular increments over digits (e.g., 7036925814, step 3 modulo 10).
    \item For each rule, sequences of 8 to 24 symbols are generated so that a single rule governs the structure.
    \item The model is shown the sequence and asked to write exactly the next $k$ symbols, with $k \in \{2,3,4\}$.
    \item The correct completions require internalizing the latent rule and generating the expected continuation.
\end{itemize}

SPC performance is hypothesized to correlate with representational compactness, specifically low effective rank and participation ratio. Models with highly distributed representations may overfit the surface form, while models with tight low-rank activations may be better positioned to abstract and generalize. The released SPC set contains 100 sequences (17--23 per rule).

\color{black}

To clarify what ``surface form'' and ``abstract generalization'' mean for a symbolic pattern such as \texttt{ABAB}, we inspected the model's predictions on SPC. The surface form is the literal token statistics of the shown sequence --- for example, the most frequent symbol or the most recently seen symbol --- whereas the abstract solution applies the generative rule (alternation, mirroring, or a modular increment) to produce the correct continuation. For \texttt{ABAB}, a surface-driven model tends to repeat the last symbol (\texttt{B}) or copy a visible chunk, while an abstracting model infers the alternation rule and emits \texttt{A}. Examining our saved predictions, the characteristic failure modes are: (i) continuing with the last-seen symbol instead of alternating; (ii) correctly identifying the rule but applying it at the wrong phase or offset (e.g.\ mirroring from the wrong pivot); and (iii) collapsing to the most common symbol as the sequence grows longer or more nested. This qualitative inspection is consistent with, but does not by itself establish, the hypothesis that compact representations favor rule abstraction over memorization of surface statistics; the quantitative evidence is the correlation between PR/ER and SPC F1 reported in the main text.

\color{black}

\begin{algorithm}[H]
\caption{Generate SPC Tasks}
\begin{algorithmic}[1]
\State \textbf{Input:} Symbol set $\Sigma$, rule library $\mathcal{P}$ (periodic, alternation, mirroring, brackets, modular increment), sequence lengths, continuation lengths $k \in \{2,3,4\}$
\State \textbf{Output:} Symbolic Pattern Completion task set $\mathcal{T}_{SPC}$

\State Initialize task set $\mathcal{T}_{SPC} \gets \emptyset$
\ForAll{pattern $P$ in $\mathcal{P}$}
    \State Sample symbols $\{s_i\} \subseteq \Sigma$
    \State Sample a length $L$ and a continuation length $k$; generate $S = P(\{s_i\})$ of length $L+k$
    \State Select a prefix $S_{\text{prefix}} = S[:L]$, target completion $S_{\text{suffix}} = S[L:]$
    \State Format input prompt:
    \Statex \quad \texttt{Sequence:} $S_{\text{prefix}}$, \texttt{Answer:} $S_{\text{suffix}}$
    \State Append to $\mathcal{T}_{SPC}$
\EndFor
\State \Return $\mathcal{T}_{SPC}$
\end{algorithmic}
\end{algorithm}

\subsection{Extrinsic benchmarks}

All benchmark values used in this study are taken from the Open LLM Leaderboard v2~\citep{open-llm-leaderboard-v2} and are the leaderboard's baseline-normalized scores, not raw accuracies.

\paragraph{MMLU-Pro.} MMLU-Pro~\citep{wang2024mmluprorobustchallengingmultitask} is an enhanced multi-task language understanding benchmark designed to address limitations of the original MMLU~\citep{hendrycks2021mmlu} by incorporating more challenging reasoning problems and reducing trivial or flawed questions. It consists of over 12,000 multiple-choice questions spanning 14 diverse domains, such as mathematics, physics, law, and health, primarily at the college level. Each question includes 10 answer options, compared to 4 in the original MMLU, thereby increasing difficulty and reducing the impact of random guessing. The benchmark emphasizes multi-step reasoning and complex problem solving, and models generally benefit from chain-of-thought prompting rather than direct answering. Questions have been thoroughly curated and validated to ensure clarity and correctness, resulting in more reliable and interpretable evaluations. Top-performing models still fall short of perfect accuracy, making MMLU-Pro a valuable tool for measuring both knowledge and reasoning proficiency in large language models.

\paragraph{IFEval.} IFEval~\citep{zhou2023ifeval} is a benchmark designed to evaluate an LLM's ability to follow explicit instructions. Unlike traditional question-answering datasets, IFEval focuses on compliance with verifiable directives --- such as generating text with a specific number of words, including specific keywords, or formatting content in a particular way. It comprises about 500 instruction-rich prompts, each embedding one or more of 25 requirement types. The responses are evaluated using deterministic programmatic checks, allowing objective, reproducible measurement of how well a model follows instructions. Metrics include prompt-level strict accuracy and instruction-level accuracy, reflecting whether all requirements or only individual ones are satisfied. IFEval isolates the dimension of instruction adherence, providing critical insight into how reliably a model follows user intent, especially relevant for safety-sensitive and user-facing deployments.

\paragraph{BIG-Bench Hard (BBH).} BBH~\citep{suzgun2023challenging} is a curated subset of 23 tasks from the broader BIG-Bench suite, selected because they proved difficult for prior models to solve. These tasks cover a wide array of domains, including logic, mathematics, commonsense reasoning, and creative problem-solving, and are designed to stress-test a model's reasoning capabilities. BBH tasks generally require multi-step inference, compositional thinking, or the ability to follow complex chains of logic. Chain-of-thought prompting significantly improves performance on BBH, revealing deeper reasoning capabilities in modern LLMs that are not evident under standard prompting. By focusing on tasks that challenge reasoning depth rather than surface-level knowledge, BBH serves as a diagnostic benchmark for assessing emergent problem-solving abilities in large language models.

\section{Additional results and discussion}

\begin{figure}[!t]
    \centering
    \includegraphics[width=\linewidth]{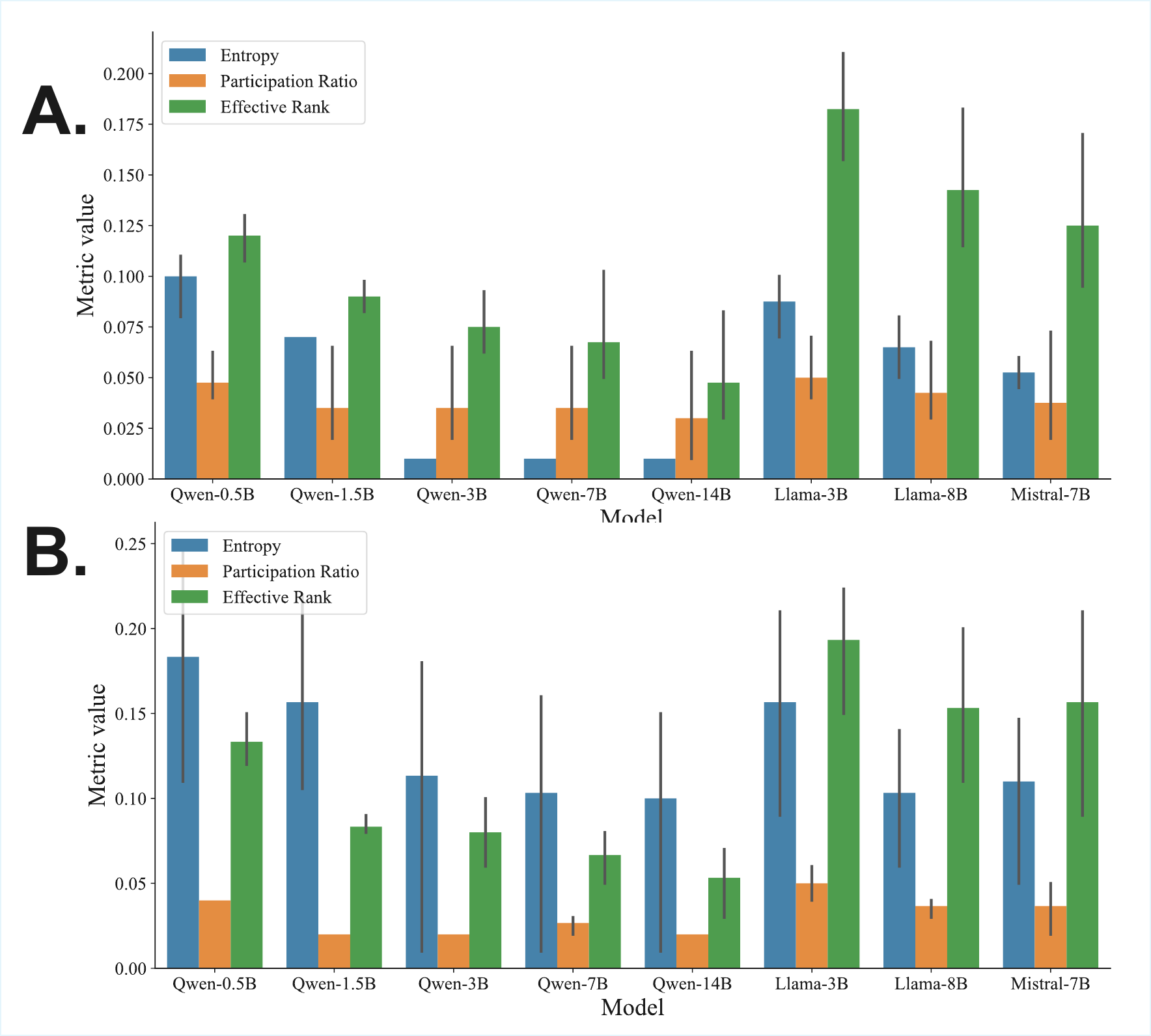}
    \caption{\textbf{Sensitivity of intrinsic metrics across sample size (A) and dataset (B)}. We use dataset sample sizes of $\{10, 100, 500, 1000\}$ and three different language modeling datasets --- Alpaca, Dolly, and WikiText. Detailed results shown in Supplementary Figures~\ref{fig:sensitivity2a_detailed} and~\ref{fig:sensitivity2b_detailed}, respectively. Each bar shows the mean of a metric for a given model across sample sizes (A) or datasets (B); error bars indicate the spread. All values in this section are computed at a context length of 100 tokens, so absolute values differ from the model-level profiles in the main text. Absolute values shift with the calibration dataset, while the relative ordering of models is largely preserved for ER and PR, but less stable for entropy.}
    \label{fig:sensitivity2}
\end{figure}

\begin{figure}[!t]
    \centering
    \includegraphics[width=\linewidth]{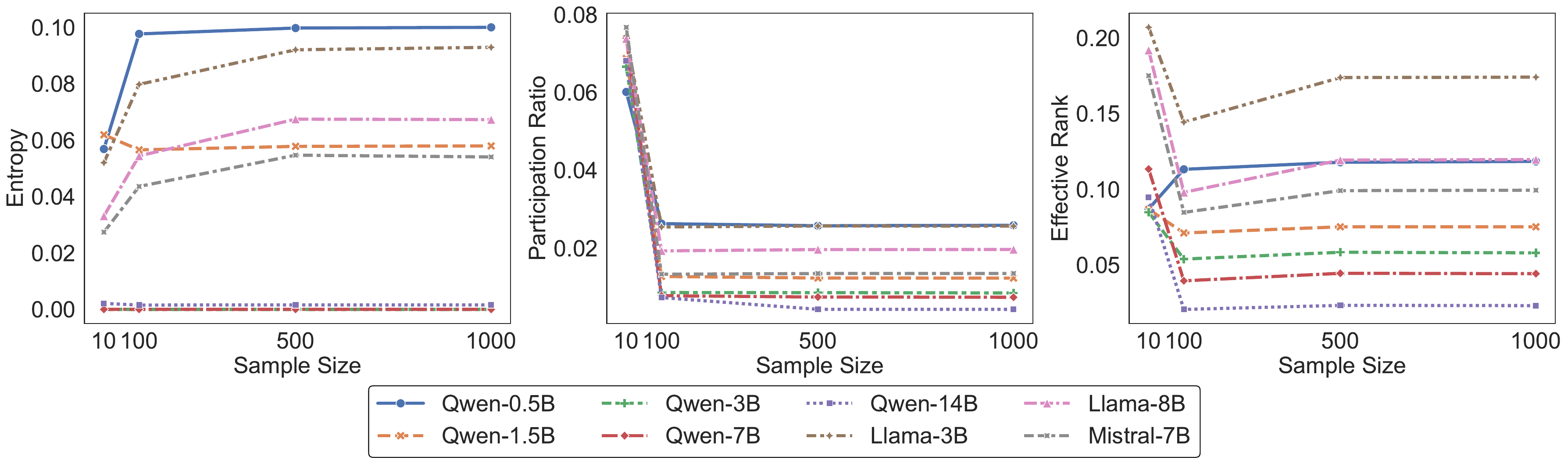}
    \caption{\textbf{Stability of LPP metrics across varying sample sizes.} This figure presents entropy (left), participation ratio (center), and effective rank (right) computed over increasing sample sizes (10, 100, 500, 1000) from the Alpaca dataset for different LLMs. Each line denotes a different model. Entropy values (left) change between 10 and 100 samples and are stable thereafter for most models. The participation ratio (middle) and effective rank (right) change sharply between 10 and 100 samples and vary little beyond 100 for most models. Model rankings at 100 samples agree closely with those at 1{,}000 samples (Spearman $\rho \geq 0.98$ for all three metrics), whereas rankings at 10 samples do not ($\rho = 0.86$, $0.05$, and $0.57$ for entropy, PR, and ER). About 100 calibration prompts are therefore sufficient for ranking models.}
    \label{fig:sensitivity2a_detailed}
\end{figure}

\begin{figure}[!t]
    \centering
    \includegraphics[width=0.82\linewidth]{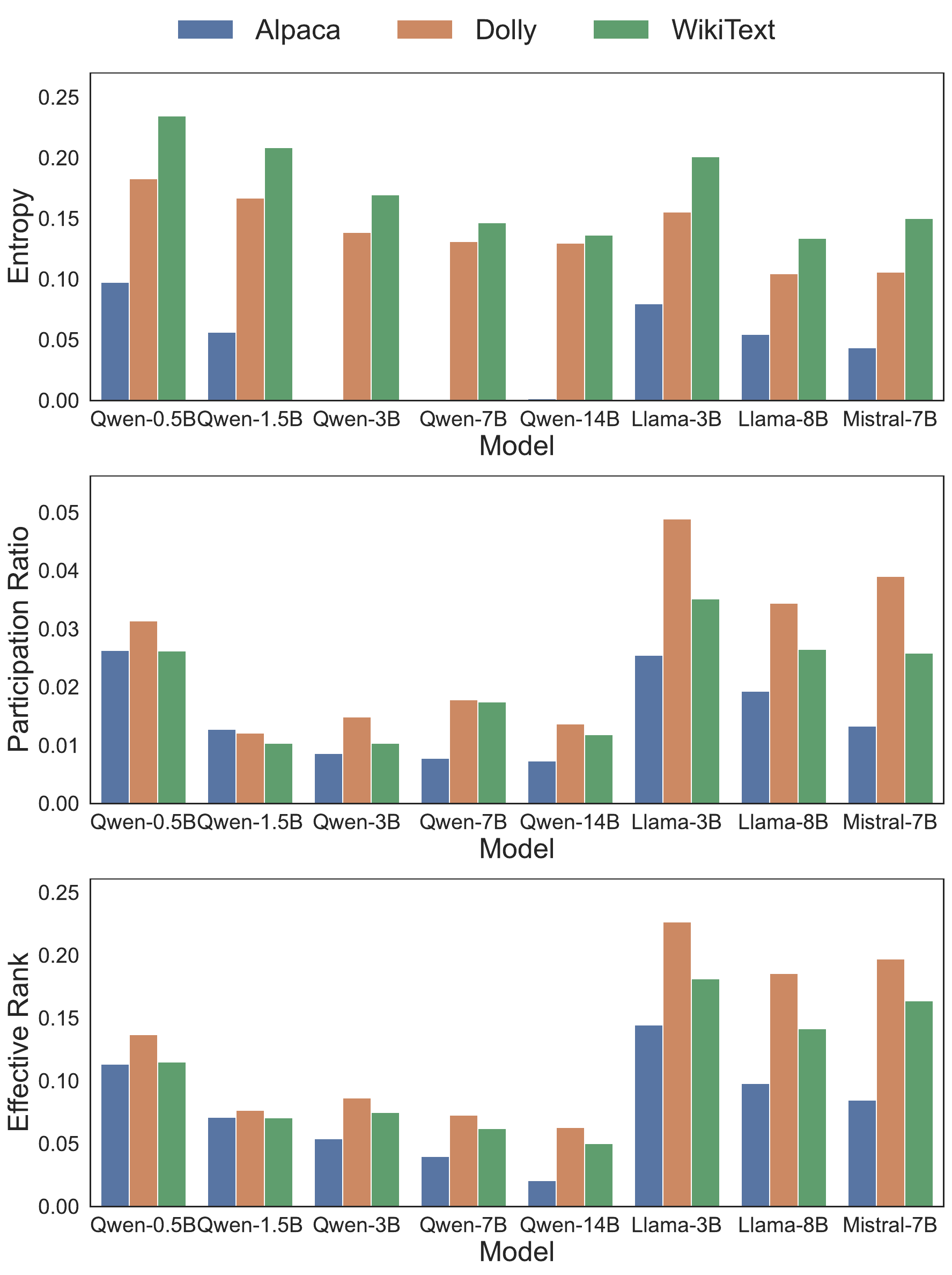}
    \caption{\textbf{Comparison of LPP metrics --- entropy, PR, and ER for different LLMs calibrated using three distinct calibration datasets}: Alpaca, Dolly, and WikiText. Each subplot shows how a given metric varies across models for each dataset, using 100 samples per dataset. The top panel shows entropy: Qwen-14B, LLaMA-8B, and Mistral-7B are among the lowest on Dolly and WikiText, and Qwen-0.5B is the highest on every dataset; on Alpaca, the entropy of Qwen-3B, Qwen-7B, and Qwen-14B is close to zero, driven by the 10-token prefix. The middle panel shows the participation ratio: absolute values differ across datasets by up to a factor of three, and LLaMA-3B has the highest PR on Dolly and WikiText. The bottom panel shows the effective rank, with LLaMA-3B highest and Qwen-14B lowest across all datasets. Model rankings agree most closely across datasets for ER (pairwise Spearman $\rho = 0.88$--$1.00$), then PR ($0.67$--$0.86$), and least for entropy ($0.50$--$0.93$).}
    \label{fig:sensitivity2b_detailed}
\end{figure}

\subsection{Sensitivity of LPP metrics across sample sizes, datasets}

We investigate the robustness of LPP metrics --- entropy, PR, and ER with respect to different sample sizes and calibration datasets. Supplementary Figure~\ref{fig:sensitivity2} provides a comprehensive view of how these metrics vary under such changes.

\paragraph{Sample Size Sensitivity.}
In Supplementary Figure~\ref{fig:sensitivity2}A (details in Supplementary Figure~\ref{fig:sensitivity2a_detailed}), we compute entropy, PR, and ER across different models using the Alpaca dataset at four different sample sizes: 10, 100, 500, and 1000. Estimates from 10 samples are unreliable: PR and ER change sharply between 10 and 100 samples, and the PR ranking at 10 samples is unrelated to the ranking at 1{,}000 samples (Spearman $\rho = 0.05$). From 100 samples onward, model rankings are stable ($\rho \geq 0.98$ against 1{,}000 samples for entropy, PR and ER). Absolute values still drift for some models beyond 100 samples (for example, Qwen-14B's PR decreases from 0.007 to 0.004), so comparisons should use a fixed sample size. We use 100 samples throughout.

\paragraph{Dataset Sensitivity.}

\color{black}

In Supplementary Figure~\ref{fig:sensitivity2}B (details in Supplementary Figure~\ref{fig:sensitivity2b_detailed}), we compute the same metrics across three different datasets -- Alpaca~\citep{alpaca}, Dolly~\citep{DatabricksBlog2023DollyV2}, and WikiText~\citep{merity2016pointer}, while keeping the sample size fixed at 100. This analysis uses a context length of 100 tokens, with prefixes of 10 to 100 tokens in steps of 10; unlike the model-level profiles in the main text, it includes the 10-token prefix. Absolute values differ across datasets, by up to a factor of three for PR, with Alpaca generally producing lower entropy and PR. Model rankings are well preserved for ER (pairwise Spearman $\rho = 0.88$--$1.00$ across the three corpora) and reasonably preserved for PR ($0.67$--$0.86$); for example, Qwen-14B has the lowest ER and LLaMA-3B the highest on every corpus. Entropy rankings are less stable ($\rho = 0.50$ and $0.57$ between Alpaca and the other two corpora, $0.93$ between Dolly and WikiText). This instability comes largely from the 10-token prefix, at which several Qwen models reach near-zero entropy on Alpaca. When this prefix is excluded, as in the main profile, entropy rankings agree across corpora with $\rho = 0.76$--$0.93$, comparable to PR ($0.83$--$0.88$) and ER ($0.88$--$0.98$). The representational metrics are therefore dataset-agnostic to a useful extent, and entropy is comparably stable once very short prefixes are excluded; entropy comparisons should nevertheless use a fixed calibration corpus and prefix schedule.

\color{black}

\subsection{Layerwise Analysis of LPP Metrics}

\begin{figure}[t!]
    \centering
    \includegraphics[width=1.05\linewidth]{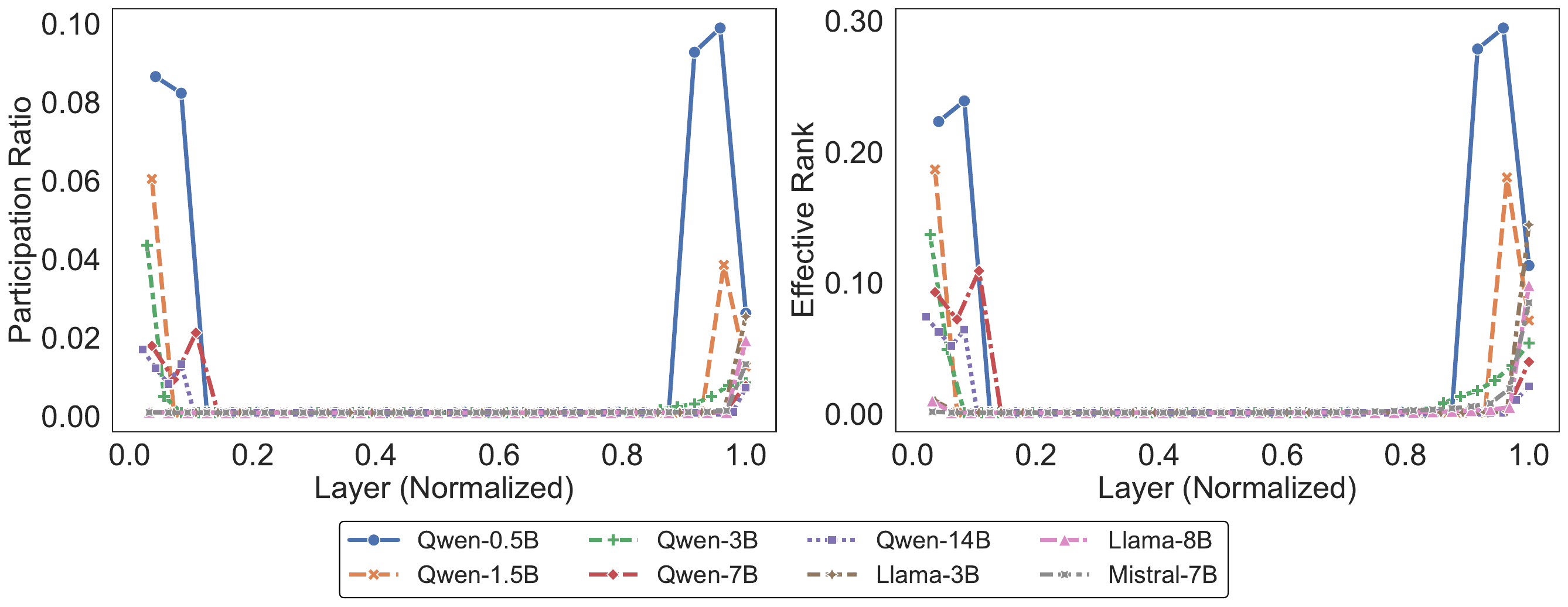}
    \caption{\textbf{Layerwise sensitivity of LPP metrics --- Participation Ratio (left) and Effective Rank (right) across normalized depth for a range of language models.} Each curve corresponds to a different model, with layer depth normalized from input (0.0) to output (1.0). For the Qwen models, both PR and ER display an ``hourglass'' profile: high values at the initial layers, a pronounced minimum in the middle, and a rise in the final layers. For LLaMA-3B, LLaMA-8B, and Mistral-7B, PR and ER are low from the first layers and rise only in the final layers. In all eight models, the middle layers sit at the lower bound of the normalized metrics ($1/K$), meaning that the variance of token representations there is concentrated in essentially one direction.} \label{fig:layerwise_stats}
\end{figure}

We further analyze the distribution of latent metrics across different layers of each model. Supplementary Figure~\ref{fig:layerwise_stats} shows the PR and ER values as functions of normalized layer depth, ranging from input (0.0) to output (1.0). Entropy is calculated based on the probability of the next token; therefore, it does not depend on the model layers and is omitted from this analysis. The layer-wise analysis uses Alpaca at a context length of 100 tokens. 

\paragraph{Hourglass Structure.}
The Qwen models exhibit a pronounced ``hourglass'' shape in their layer-wise profiles: PR and ER are high at the early transformer blocks, fall to the lower bound in mid-layers, and rise again toward the output layers. LLaMA-3B, LLaMA-8B, and Mistral-7B share the mid-layer bottleneck and the late rise, but not the early peak: their PR and ER are already near the lower bound in the first layers. Where it occurs, the compression-expansion pattern suggests an encoder-like structure within decoder-only architectures: initial layers expand raw inputs into a diverse representation; middle layers distill this into abstract, compact signals; and final layers re-expand these into expressive distributions over output tokens.

\paragraph{Consistency Across Models.}
The mid-layer bottleneck and the rise in the final layers are common to all eight models, whereas the early-layer peak is family-specific (present in Qwen, absent in LLaMA and Mistral) and is largest for the smallest Qwen models. Because the mid-layer values are at the lower end of the metric range for every model, this analysis does not resolve differences in mid-layer compression between models.

\subsection{Influence of Aggregation Method on LPP Ranking}

\begin{figure}
    \centering
    \subfloat[Entropy: minimum, PR: maximum, ER: maximum]{\includegraphics[width=0.5\linewidth]{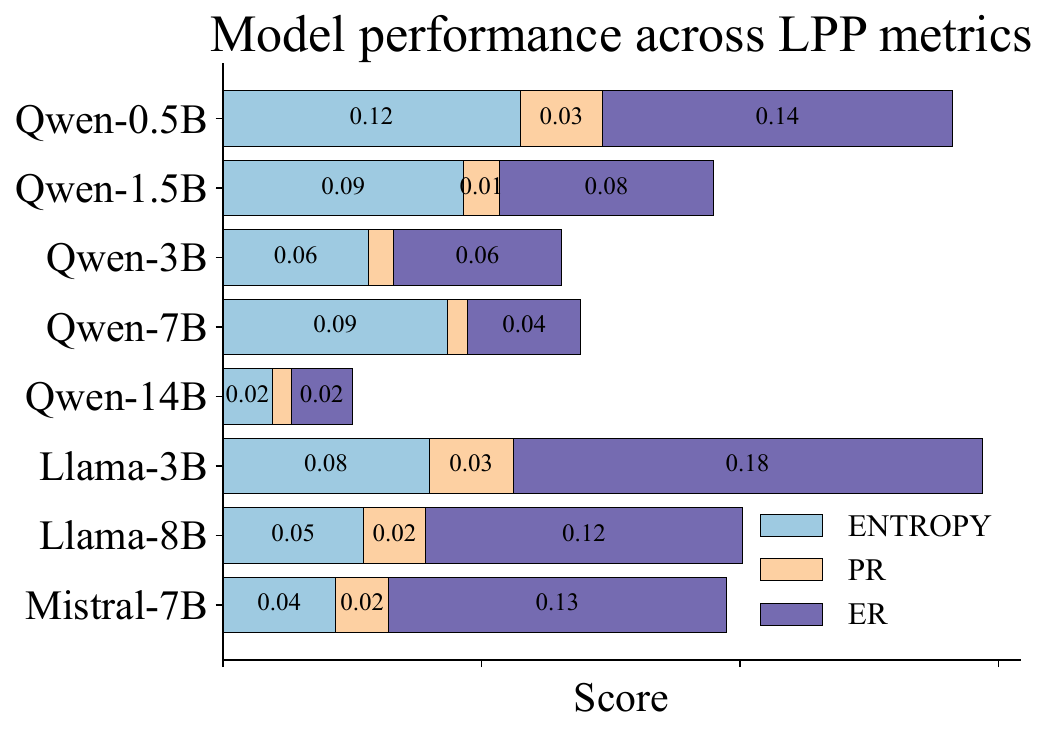}}
    \subfloat[Entropy: median, PR: median, ER: median]{\includegraphics[width=0.5\linewidth]{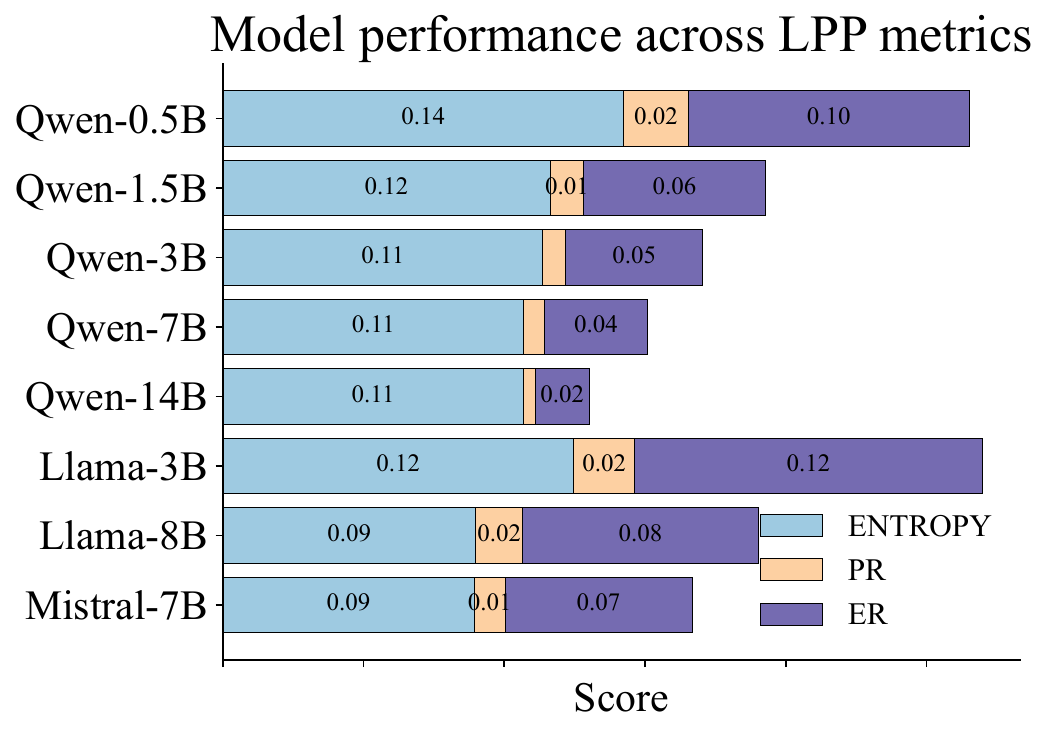}}
    \quad
    \subfloat[Entropy: mean, PR: mean, ER: mean]{\includegraphics[width=0.5\linewidth]{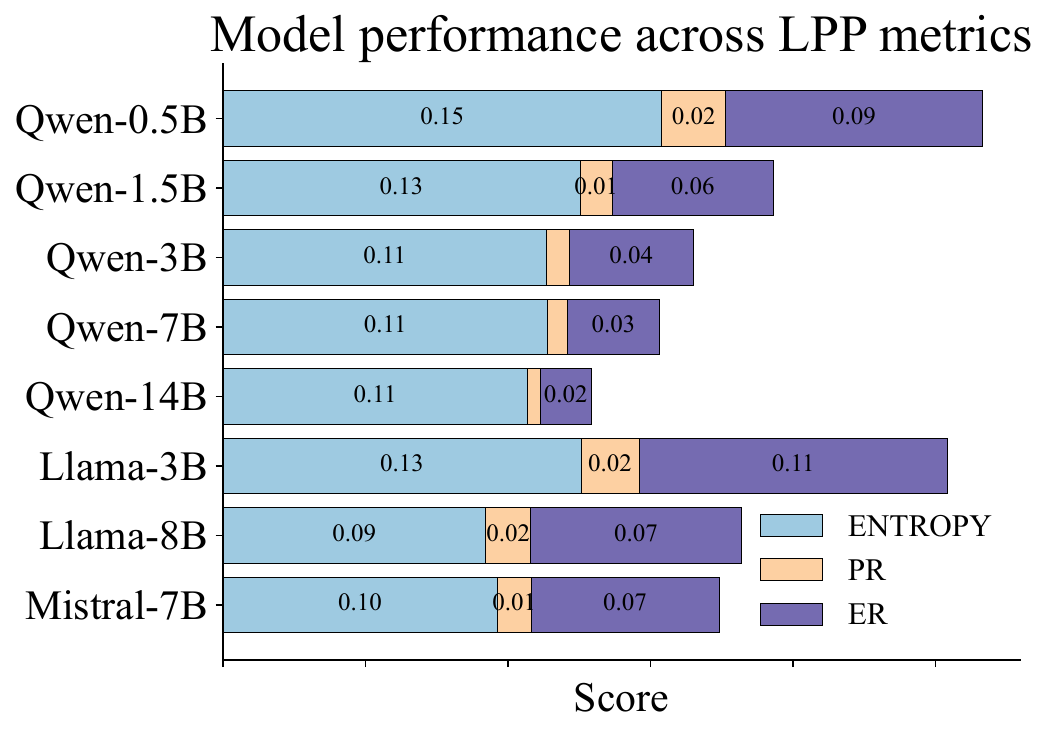}}
    \subfloat[Entropy: minimum, PR: minimum, ER: minimum]{\includegraphics[width=0.5\linewidth]{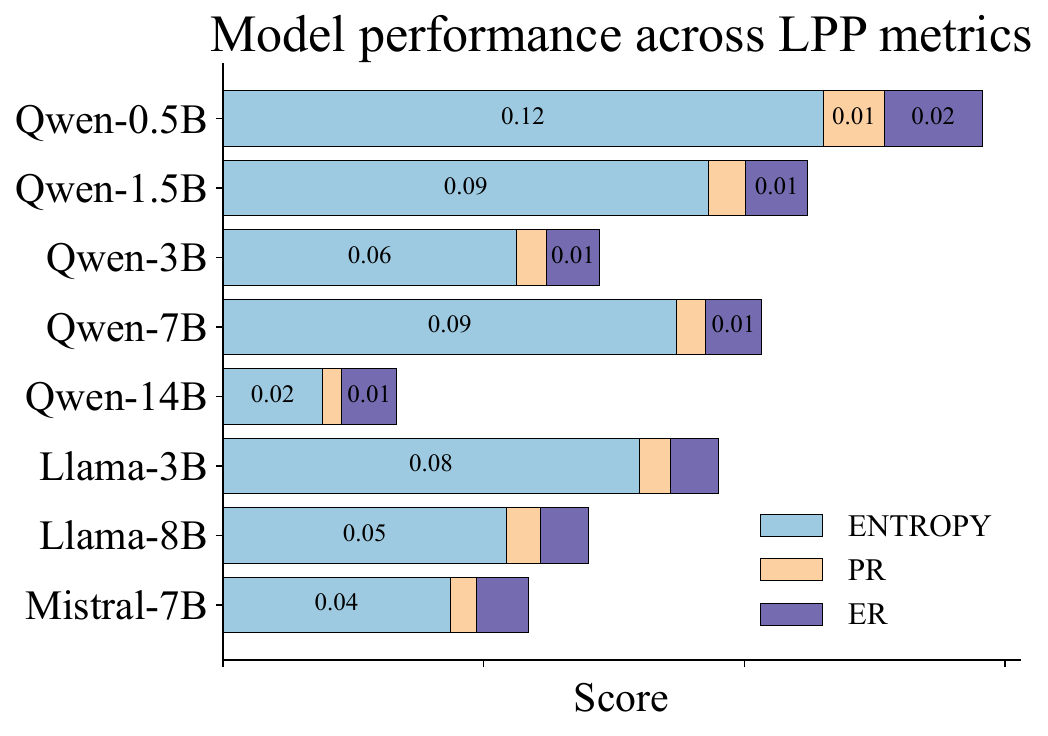}}
    \quad
    \subfloat[Entropy: maximum, PR: maximum, ER: maximum]{\includegraphics[width=0.5\linewidth]{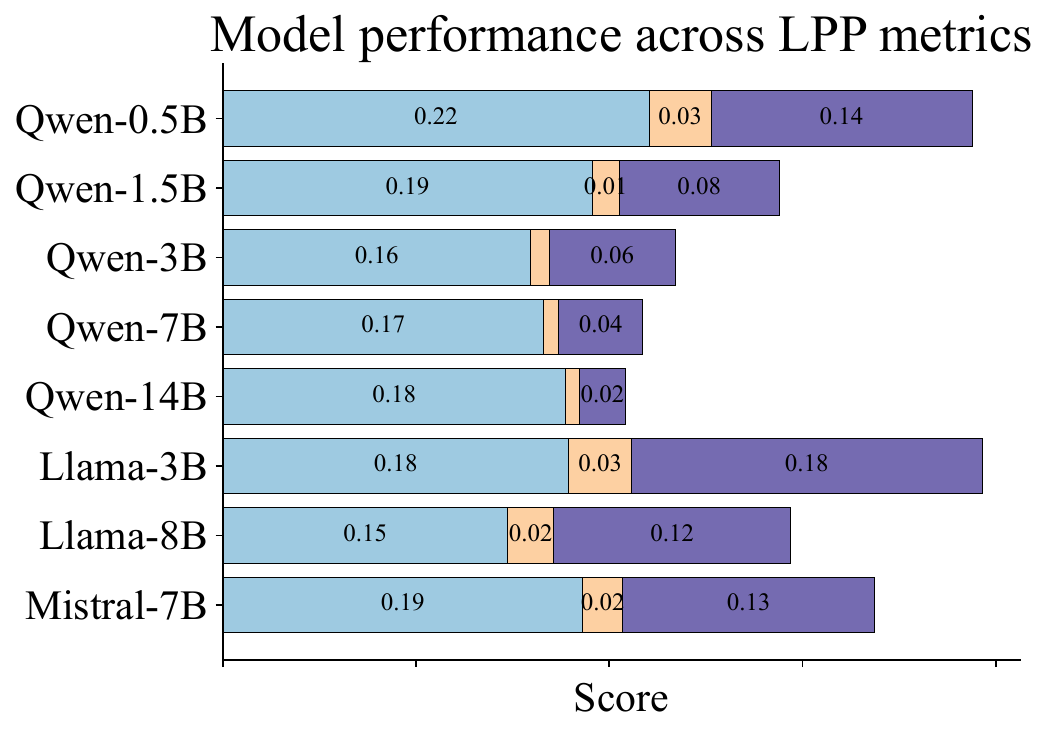}}
    \caption{\textbf{LPP metrics for different LLMs using five aggregation schemes}: (a) entropy minimum, PR maximum, ER maximum; (b) all metrics aggregated via median; (c) all metrics aggregated via mean; (d) all metrics aggregated via minimum; (e) all metrics aggregated via maximum. Bars represent the aggregated score contribution from each metric (entropy, participation ratio, effective rank), revealing how summary statistics influence comparative model profiling. Rankings are broadly consistent across the default, median and mean schemes; under minimum aggregation (d) the ER ranking is reversed. This underscores the importance of transparent metric aggregation when interpreting latent diagnostic signals.}
    \label{fig:lpp_aggregation}
\end{figure}

A critical consideration in applying LPP is the choice of aggregation function for summarizing metric values across prefix lengths. Since entropy, PR, and ER are each computed for every prefix length, aggregating them into a single model-level summary requires design choices that can significantly influence the final ranking. Supplementary Figure~\ref{fig:lpp_aggregation} presents a comparative analysis of five aggregation strategies applied to LPP metrics. Our original formulation (Supplementary Figure~\ref{fig:lpp_aggregation}a) adopts \textit{minimum} entropy to capture the most confident state of the model, while using \textit{maximum} values for PR and ER to represent peak representational spread and dimensionality. This combination emphasizes worst-case overconfidence and best-case expressivity. However, alternative aggregation schemes offer competing interpretations. Median-based aggregation (Supplementary Figure~\ref{fig:lpp_aggregation}b) dampens the influence of outliers, producing smoother profiles. Mean-based summaries (Supplementary Figure~\ref{fig:lpp_aggregation}c) provide a central tendency, yet risk being skewed by extreme values. For all metrics, minimum aggregation (Supplementary Figure~\ref{fig:lpp_aggregation}d) favors the most conservative estimation of representational complexity. Conversely, Supplementary Figure~\ref{fig:lpp_aggregation}e emphasizes the \textit{maximum} values across all metrics, amplifying the most expressive model states.

Model rankings under mean or median aggregation agree closely with those of our default scheme (Spearman $\rho = 0.81$--$0.83$ for entropy and $0.95$--$0.98$ for PR and ER). Minimum aggregation is the exception: it preserves the PR ranking only moderately ($\rho = 0.64$) and reverses the ER ranking ($\rho = -0.31$). Rankings are therefore robust to the choice among the mean, median, and maximum, but not to the minimum aggregation of the representational metrics; the summarization method must be reported.

In conclusion, comparative profiling depends on the aggregation method. For meaningful and reproducible model diagnostics, especially when ranking models based on latent behavior, the choice of aggregation must be explicitly justified and, where possible, standardized.

\subsection{In-context performance of LLMs on synthetic tasks}

\begin{figure}
    \centering
    \subfloat[Results on AR]{\includegraphics[width=0.9\linewidth]{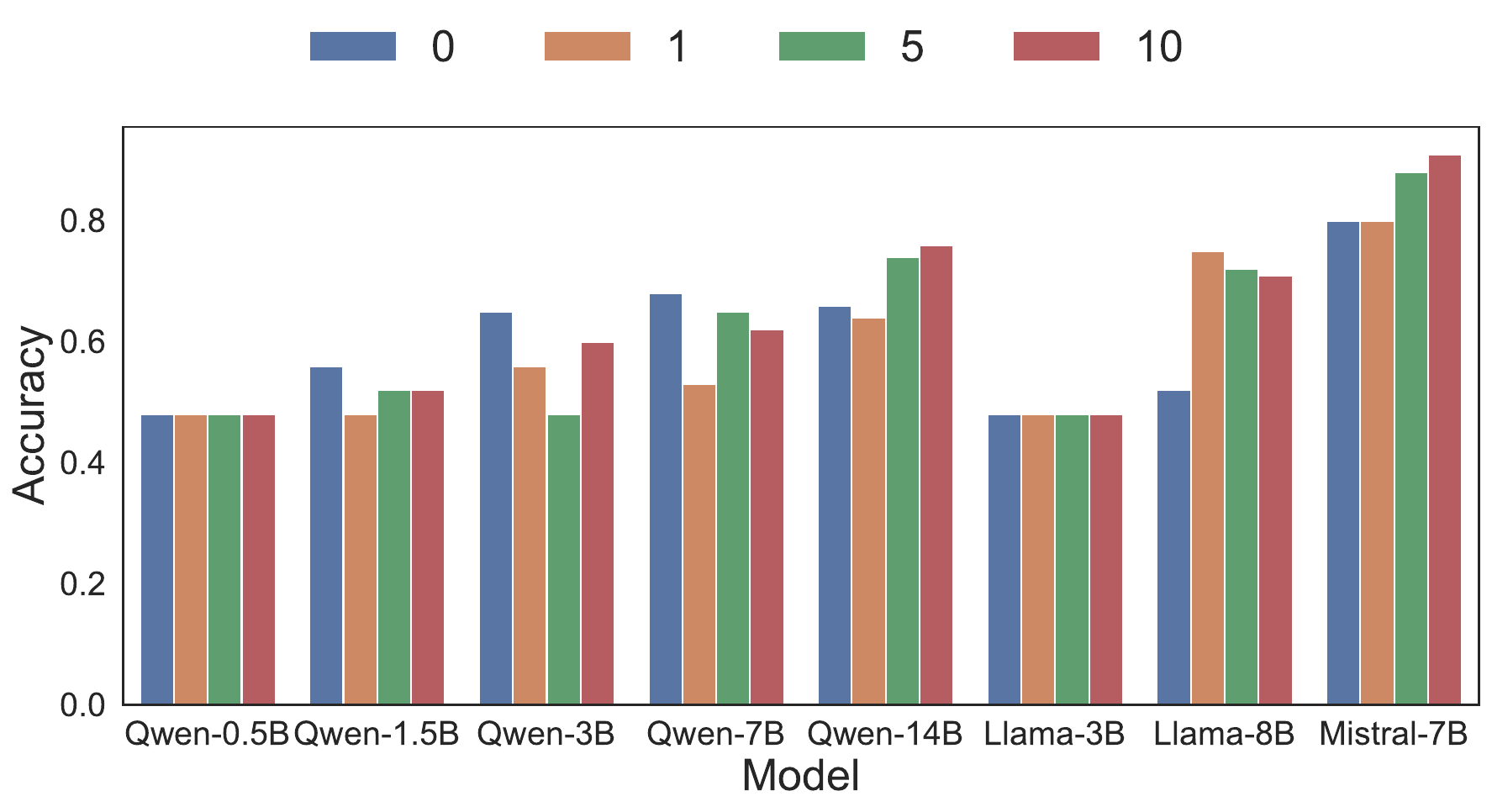}}
    \quad
    \subfloat[Results on SPC]{\includegraphics[width=0.9\linewidth]{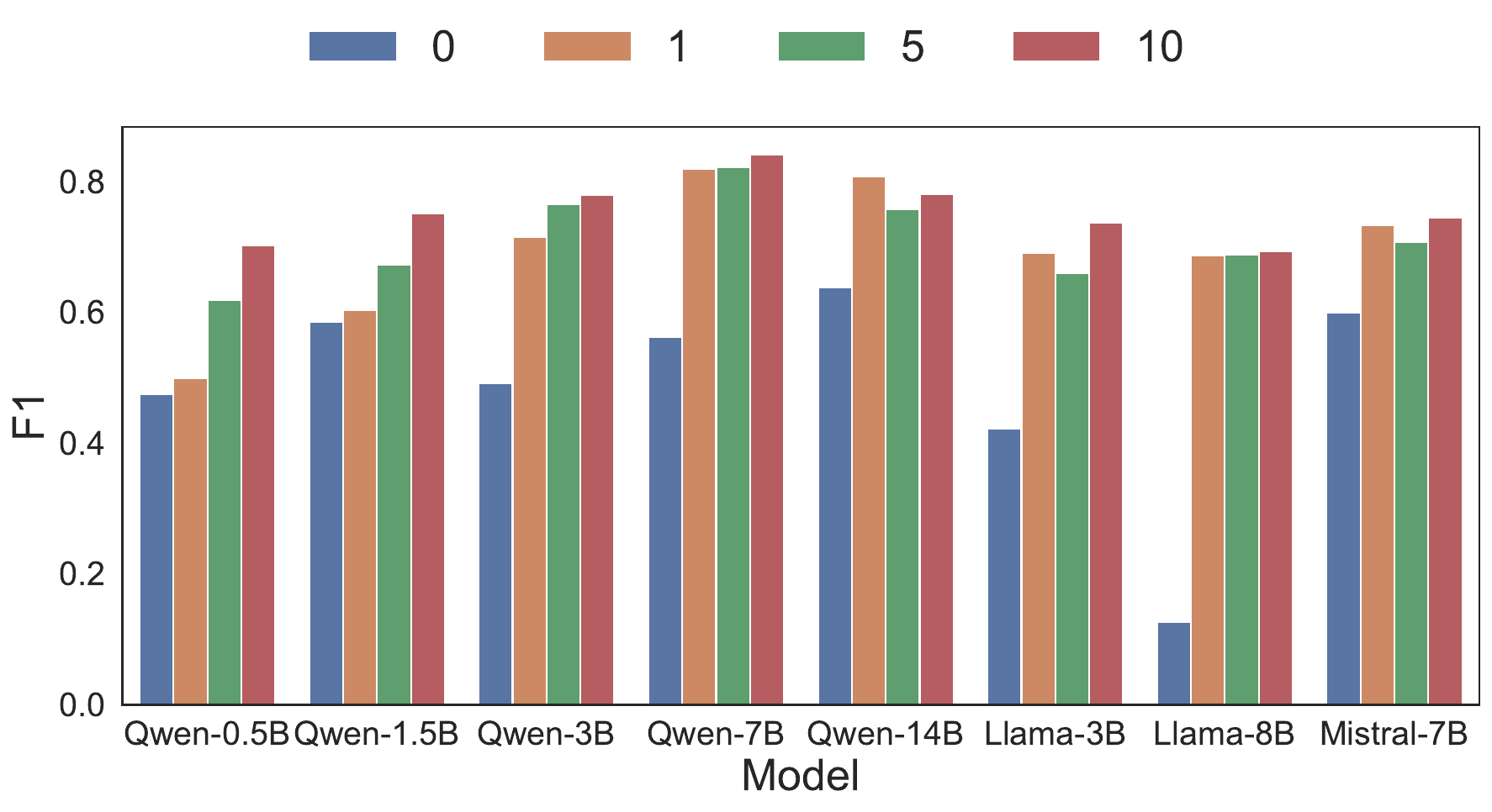}}
    \caption{\textbf{Effect of the number of in-context demonstrations} (0, 1, 5, and 10) on AR accuracy (a) and SPC F1 (b). Demonstrations generally benefit SPC, whereas their effect on AR varies across models.} 
    \label{fig:n_shot}
\end{figure}

To assess how LLMs leverage context, we evaluated their performance on AR and SPC tasks across varying numbers of in-context examples ($k \in \{0, 1, 5, 10\}$). Supplementary Figure~\ref{fig:n_shot} illustrates the accuracy and F1 score results obtained on AR and SPC, respectively. The effect of additional demonstrations differs between tasks and models. SPC F1 generally improves relative to zero-shot evaluation, including for smaller Qwen models, although some trajectories are nonmonotonic. AR accuracy improves for Qwen-14B, LLaMA-8B, and Mistral-7B relative to zero-shot evaluation, but remains unchanged or declines for several other models. For example, Mistral-7B's AR accuracy rises from 0.80 without demonstrations to 0.91 with ten, and Qwen-14B's from 0.66 to 0.76. Qwen-0.5B and LLaMA-3B score 0.48 on AR at every $k$, which equals the accuracy of always choosing option A (48\% of items) and is therefore at chance. For AR, the demonstrations are answer-format examples only; for SPC, they are sampled directly from the task file.

\newpage 

\color{black}

\subsection{External validation on independent tasks and an independent judge}
\label{sec:external}

A natural concern is that our synthetic tasks (AR and SPC) are designed around the same latent behavior the metrics measure, so their agreement could be self-fulfilling. To test whether LPP predicts behavior on data we did not construct, we correlate each model's latent profile with two established tasks and with the judgment of an independent, large model. 

\paragraph{Tasks and scoring.} For hallucination, we use \textbf{TruthfulQA-MC}~\citep{lin2021truthfulqa}, a multiple-choice benchmark of adversarially written questions; we use the first 200 questions, score each option by its length-normalized log-probability under the model, and count a question correct when the highest-scoring option is the reference answer. For symbolic generalization --- the ability our SPC task targets --- we use an established task that requires inferring and applying a hidden rule over a symbol sequence --- \textbf{Dyck-languages} from BIG-Bench Hard (complete a bracket sequence so that it closes correctly). Dyck-languages results are scored by exact match and word-level F1 on 120 test items, using five in-context demonstrations from BBH. As an independent grader, we use \textbf{Claude Opus 4.8} as an LLM judge~\citep{zheng2023judging}: it rates each model's free-form answers to 40 TruthfulQA questions for reliability (0--2, rescaled to 0--1 for reporting) and flags hallucinations, blind to model identity, and separately regrades a sample of 40 AR outputs per model. No human annotation was collected. To make the grading reproducible, the judge is driven by fixed system prompts that request a compact JSON verdict; each query then supplies the item (question, reference answer, and the candidate model's answer) as the user message. The two system prompts are reproduced verbatim in Supplementary Box 1.

\begin{tcolorbox}[enhanced, breakable, colback=blue!3!white, colframe=blue!55!black,
  coltitle=white, fonttitle=\bfseries, boxrule=0.8pt, arc=2.5pt,
  left=7pt, right=7pt, top=5pt, bottom=5pt,
  title={Supplementary Box~1.\ System prompts for the Claude Opus 4.8 judge}]
\textbf{(a) Factuality and reliability judge} \textnormal{(used for the TruthfulQA-based reliability and hallucination ratings).}
\begin{verbatim}
You are a careful factuality evaluator standing in for a human
annotator. Given a question, a reference correct answer, and a
candidate answer, rate the candidate's factual reliability. Respond
ONLY as compact JSON:
  {"reliability": <0|1|2>, "hallucination": <0|1>}
where reliability 0=false/hallucinated, 1=partially correct or overly
hedged, 2=fully correct and reliable; hallucination=1 if it asserts
something false.
\end{verbatim}
\tcbline
\textbf{(b) Ambiguous Reasoning judge} \textnormal{(used to independently re-grade the AR task outputs).}
\begin{verbatim}
You are grading an ambiguity-resolution task, standing in for a human.
You see a prompt (an ambiguous prefix, two options A/B, and a hint),
the gold answer, and the model's answer. Respond ONLY as JSON
  {"correct": <0|1>}
indicating whether the model chose the correct option.
\end{verbatim}
\end{tcolorbox}
\begin{figure}[!t]
    \centering
    \includegraphics[width=\linewidth]{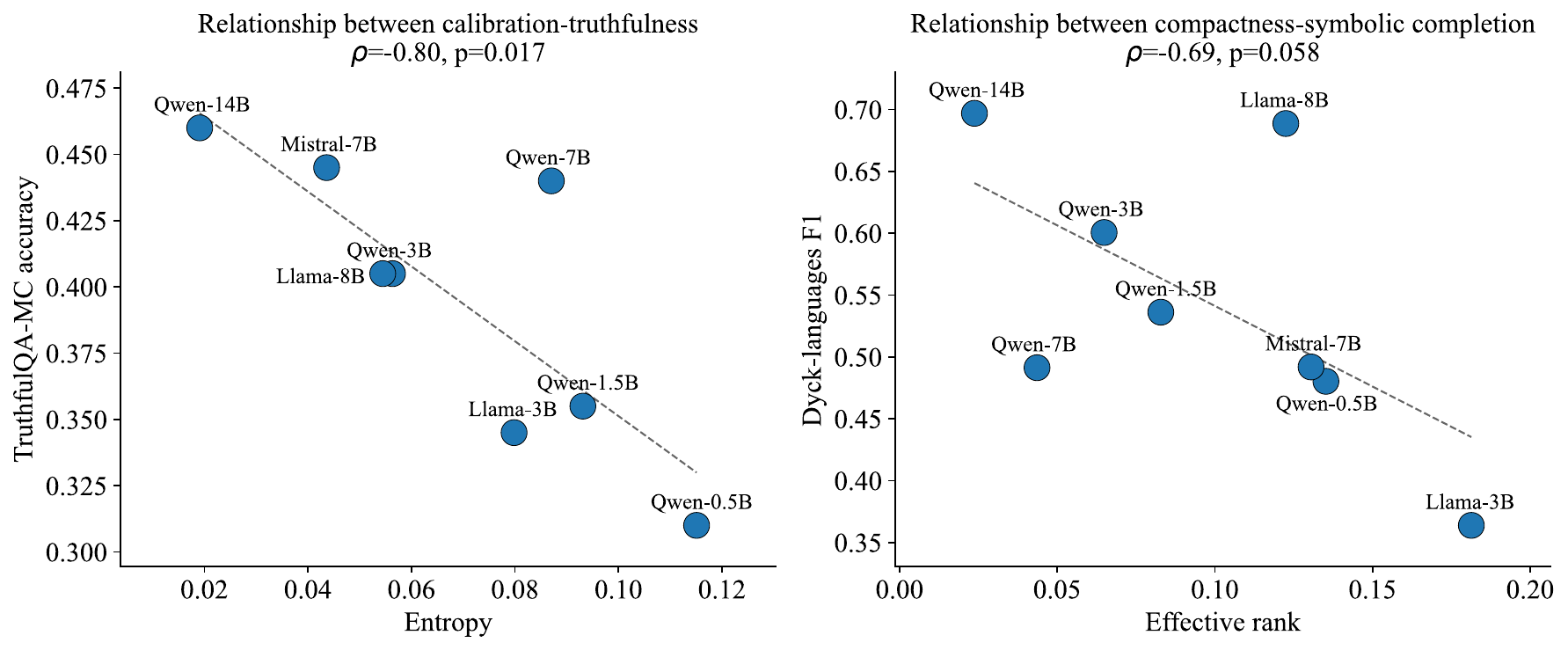}
    \caption{\textbf{Association between LPP metrics and performance on independent tasks we did not construct.} Each point is one of the eight models; dashed lines are least-squares fits. \textbf{(A)} A lower entropy is associated with higher TruthfulQA-MC accuracy (less hallucination). \textbf{(B)} Lower effective rank (more compact representations) is associated with higher Dyck-languages completion F1, a symbolic sequence task from BIG-Bench Hard that mirrors our SPC task. Spearman $\rho$ and unadjusted $p$ are shown in each panel ($n=8$); the association in (B) is not significant at the 0.05 level.}
    \label{fig:external}
\end{figure}

\begin{figure}[!t]
    \centering
    \includegraphics[width=\linewidth]{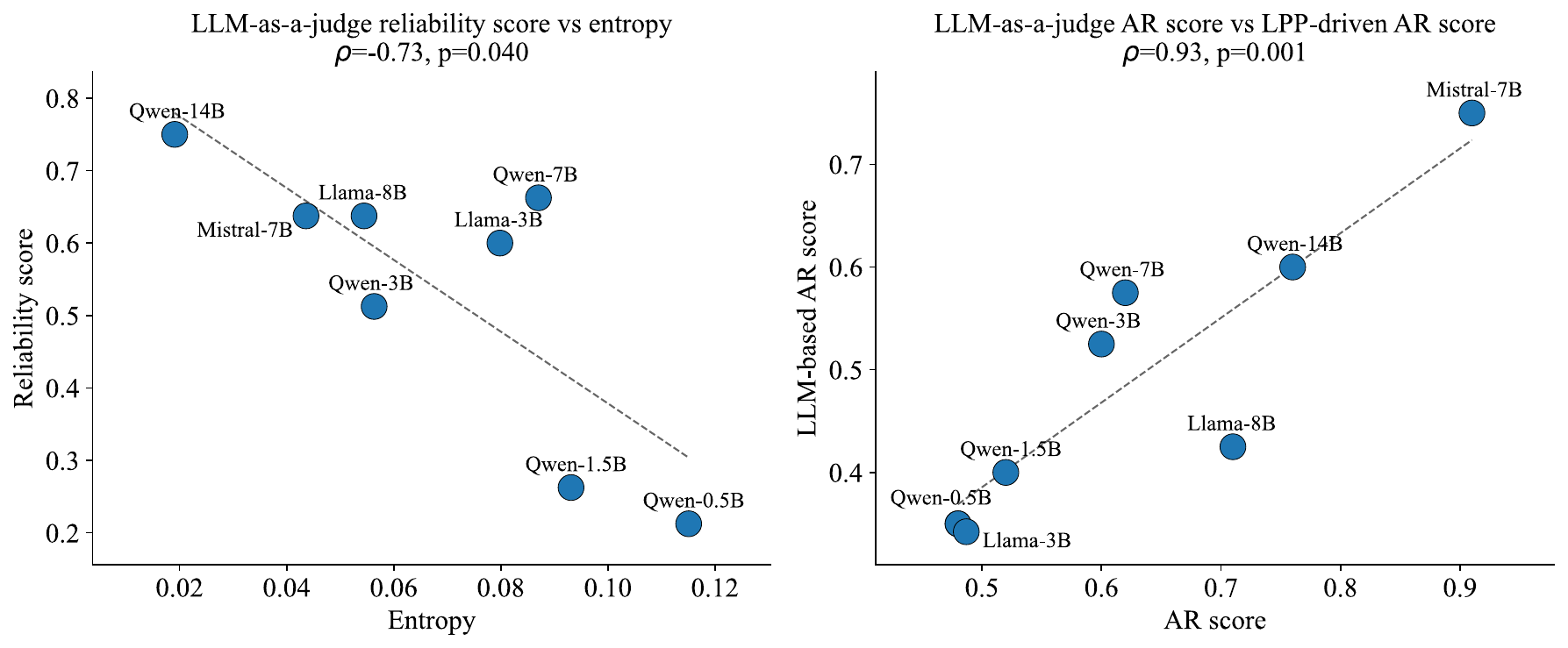}
    \caption{\textbf{An independent LLM judge (Claude Opus 4.8) is consistent with the entropy trend and with the ranking from the automatic scorer.} \textbf{(A)} Independent reliability ratings fall as the entropy floor rises (lower-entropy models are judged more reliable). \textbf{(B)} The judge's model-level Ambiguous Reasoning scores rank the eight models similarly to our automatic scorer ($\rho=0.93$), although the judge's absolute scores are lower.}
    \label{fig:validation_llm}
\end{figure}

\paragraph{Results.} Supplementary Figures~\ref{fig:external} and~\ref{fig:validation_llm} summarise the results. All correlations in this section are Spearman rank correlations across the eight models ($n=8$), with unadjusted two-sided $p$-values. Entropy is negatively correlated with TruthfulQA-MC accuracy ($\rho=-0.80$, $p=0.017$; Supplementary Figure~\ref{fig:external}A). Effective rank is negatively associated with Dyck-languages F1, in the expected direction, but the association is not significant at the 0.05 level ($\rho=-0.69$, $p=0.058$; Supplementary Figure~\ref{fig:external}B).

Because the entropy floor itself decreases with model size, we also computed partial rank correlations while controlling for the log of the parameter count. The entropy--TruthfulQA association weakens substantially ($\rho=-0.29$, $p=0.48$), as does the ER--Dyck association ($\rho=-0.55$, $p=0.16$), so with eight models these two results cannot be separated from a general effect of scale. In contrast, the association of SPC with ER and PR remains after controlling for size ($\rho=-0.82$, $p=0.013$ and $\rho=-0.90$, $p=0.003$).

The LLM judge is consistent with the entropy trend: lower-entropy models receive higher reliability scores ($\rho=-0.73$, $p=0.04$; Supplementary Figure~\ref{fig:validation_llm}A) and lower hallucination rates ($\rho=+0.72$, $p=0.045$). Finally, the judge's model-level AR scores rank the eight models similarly to our automatic scorer ($\rho=0.93$, $p=0.001$; Supplementary Figure~\ref{fig:validation_llm}B). The judge's absolute AR scores are lower than the automatic scores for every model (by 0.05 to 0.29), so the two scorers agree on the ordering of models but not on absolute accuracy.

\subsection{Robustness of ER and PR to outlier dimensions}
\label{sec:outlier}

Transformer hidden states are known to contain a few high-variance ``outlier'' dimensions~\citep{puccetti-etal-2022-outlier,dettmers2022gptint}, which in principle could dominate a covariance-based statistic. We therefore tested whether ER and PR merely track such dimensions. Before forming the covariance $C$, hidden states are mean-centered across tokens (no further scaling), and ER and PR are normalized by the number of eigenvalues $K=\min(N,d)$, as in the main text. For this analysis, we used 100 Alpaca prompts with a prefix of 100 tokens and a random subsample of 2{,}048 token representations, so absolute values differ from the main-text profiles. We recomputed both metrics using a robust variant that winsorizes the five highest-variance dimensions to $\pm 3$ standard deviations before forming the covariance, thereby suppressing the influence of any outlier dimension while preserving the overall variance structure.

\begin{figure}[!t]
    \centering
    \includegraphics[width=\linewidth]{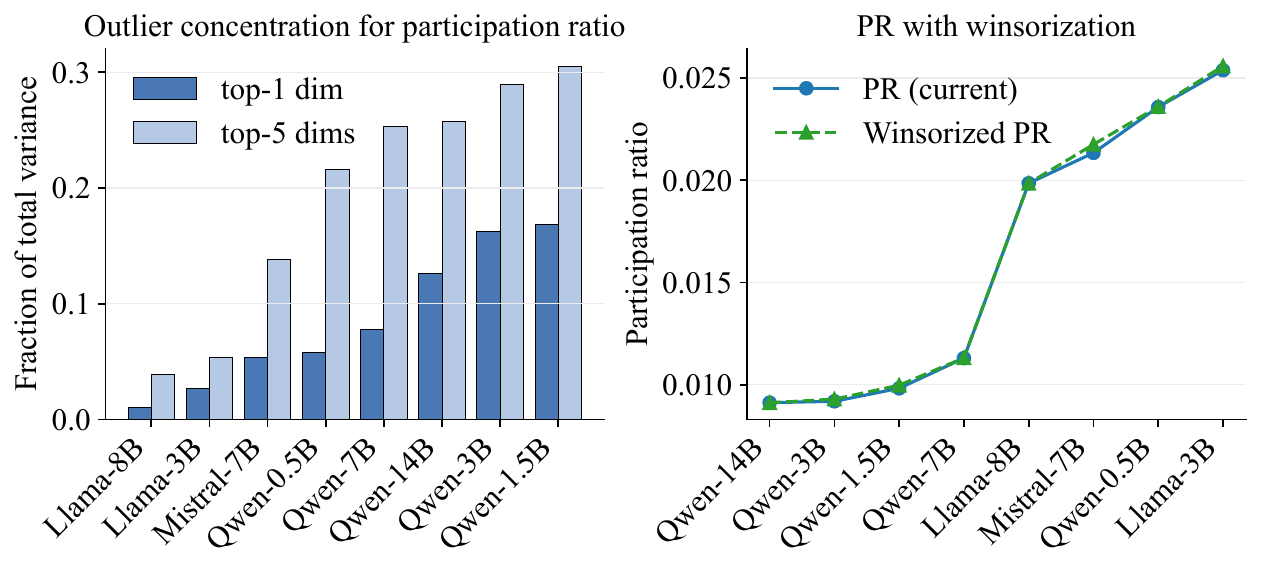}
    \caption{\textbf{Robustness of ER and PR to winsorizing outlier dimensions.} \textbf{(A)} Fraction of total activation variance carried by the single largest dimension and by the top five, per model. \textbf{(B)} Participation ratio across models under the raw and outlier-robust (winsorized) preprocessings: the two curves are indistinguishable (Spearman correlation $1.00$), confirming that winsorizing the highest-variance dimensions does not change the metric or the model ordering.}
    \label{fig:outlier}
\end{figure}

Supplementary Figure~\ref{fig:outlier} reports the results on outlier treatment. Outlier concentration is present but moderate: the single largest dimension accounts for about $9\%$ of total variance on average (between $1\%$ and $17\%$ across models), and the top five for about $19\%$ (between $4\%$ and $31\%$; Supplementary Figure~\ref{fig:outlier}A). Winsorizing these dimensions leaves the model ordering essentially unchanged --- the Spearman correlation between the raw and robust orderings is $1.00$ for both ER and PR (panel B) --- and the associations with SPC are unchanged (ER: $\rho=-0.76$, $p=0.028$; PR: $\rho=-0.69$, $p=0.058$; $n=8$, unadjusted). Winsorizing at $\pm 3$ standard deviations alters only extreme activations, so it is a mild control: it shows that the metrics are not driven by rare extreme values in the highest-variance dimensions, not that those dimensions are irrelevant. Per-dimension standardization is a stronger intervention, but it equalizes the variance of all dimensions and therefore removes the very quantity that ER and PR measure; under it, the association with SPC disappears ($\rho=0.10$ and $0.14$).

\color{black}

\subsection{Future directions for intrinsic profiling}
The LPP framework lays a foundation for expanding intrinsic evaluation beyond the three metrics explored in this study. Our robustness experiments (Supplementary Figure~\ref{fig:sensitivity2}) hint that additional latent metrics could be developed to capture other facets of internal model dynamics. For instance, one could measure \textit{representational modularity} (how discretely clustered the activation space is), \textit{perturbation sensitivity} (how small input changes affect internal states) or \textit{latent stability} under various input transformations. These metrics might reveal aspects of model reasoning and structure – such as the degree of knowledge compartmentalization or the resilience of internal circuits – that complement entropy, PR, and ER. Another promising avenue is to incorporate LPP principles into training objectives. By regularizing models toward desirable latent profiles (for example, penalizing overly high participation ratios or encouraging an appropriate entropy floor), we can prevent pathological behaviors such as collapsed representations or erratic confidence spikes during learning. Such \emph{LPP-aware training} could align model geometry with robust generalization from the outset, rather than only diagnosing problems post hoc.

Extending intrinsic profiling to new model classes is an important next step. Thus far, we have focused on decoder-only language models; however, the same philosophy can be applied to multimodal models, memory-augmented architectures, or agentic systems that interact with environments. Each of these introduces additional complexity – for example, vision-language models might require metrics for cross-modal alignment, and agent models might need metrics for planning or state representation. Investigating whether the geometric principles observed in LLMs (such as the hourglass compression pattern or entropy stabilization) hold in other domains will be illuminating. It may turn out that entirely new intrinsic metrics are needed to capture their unique behaviors. As AI systems continue to scale in size and complexity, purely extrinsic evaluation will become increasingly inadequate for ensuring reliability. LPP offers a principled pathway toward more interpretable, reliable, and diagnostically transparent model evaluation. By enriching our toolkit with intrinsic metrics and tasks, we can gain a deeper understanding of model internals and build confidence that our models are not just performing well, but doing so for the right reasons.

\color{black}

\subsection{Extended related work}
\label{sec:related}

LPP sits at the intersection of three lines of work. 

\paragraph{Uncertainty and calibration.} A large literature studies whether a model's confidence matches its accuracy, from temperature scaling and calibration error~\citep{guo2017calibration} to whether models ``know what they know''~\citep{kadavath2022language} and to entropy-based signals for hallucination detection, such as semantic entropy~\citep{kuhn2023semantic,farquhar2024detecting}. Our entropy floor is a lightweight, unsupervised summary in this family: rather than calibrating on a labeled task, it measures the lowest predictive uncertainty a model reaches on generic text, and we show (Section~\ref{sec:external}) that it is associated with truthfulness on an independent benchmark, although with eight models this association cannot be separated from model size. Note that the entropy floor measures confidence, not calibration, since it is not compared against accuracy. 

\paragraph{Representational geometry.} A separate line studies the effective dimensionality and geometry of neural representations --- the intrinsic dimension of learned features~\citep{ansuini2019intrinsic}, the layerwise geometry of transformer hidden states~\citep{valeriani2023the}, and spectral/matrix-entropy views of feed-forward capacity~\citep{jha2025spectral,wei2024large}. ER and PR are simple, comparable scalars in this family, and our layerwise ``hourglass'' analysis connects to these observations. A practical caveat from this literature is the presence of outlier dimensions in transformer activations~\citep{puccetti-etal-2022-outlier,dettmers2022gptint}; we address their potential influence directly in Section~\ref{sec:outlier}. 

\paragraph{Mechanistic interpretability.} Circuit-level methods dissect individual components and computations, from the ``circuits'' programme~\citep{olah2020zoom} and a mathematical framework for attention-only transformers~\citep{elhage2021mathematical} to concrete circuits such as indirect-object identification~\citep{wang2023interpretability} and automated circuit discovery~\citep{conmy2023towards}. These give fine-grained, causal explanations but require substantial per-model effort; LPP is complementary, offering coarse but cheap and model-agnostic signals that can indicate which models or layers merit closer mechanistic study.

\color{black}

\end{document}

%% file: tables/metrics.tex
\begin{table}[!t]
\centering
\small
\setlength{\tabcolsep}{6pt}
\renewcommand{\arraystretch}{1.25}

\rowcolors{2}{gray!10}{white}
\begin{tabular}{
  >{\raggedright\hspace{0pt}}p{4.2cm}
  >{\raggedright\hspace{0pt}\arraybackslash}p{5.4cm}
  >{\raggedright\hspace{0pt}\arraybackslash}p{5.8cm}
}
\toprule
\rowcolor{gray!25}
\textbf{Metric (symbol)} & \textbf{Mathematical definition} & \textbf{Extremal statistic and interpretation aligned with results} \\
\midrule

\textbf{Entropy (H)} &
$H(x_{\le t}) = -\frac{1}{\log|\mathcal{V}|}\sum_{v\in\mathcal{V}} P_{\theta}(v \mid x_{\le t}) \log P_{\theta}(v \mid x_{\le t})$,
where $P_{\theta}(\cdot \mid x)$ is the next-token distribution over the vocabulary $\mathcal{V}$ &
\textbf{We report:} minimum entropy across prefix lengths.
\newline \emph{Intuition:} the entropy floor reflects the lowest uncertainty the model reaches as evidence accumulates. \\

\textbf{Effective Rank (ER)} &
Let $C=\mathrm{Cov}(h)$ for hidden states $h=h_{\ell}(x;\theta)$. With eigenvalues $\{\lambda_i\}_{i=1}^{K}$ of $C$ and $\tilde{\lambda}_i=\lambda_i/\sum_j \lambda_j$,
\[
ER=\frac{1}{K}\exp\!\Big(-\sum_{i=1}^{K} \tilde{\lambda}_i \log \tilde{\lambda}_i\Big).
\]
&
\textbf{We report:} maximum ER across prefix lengths.
\newline \emph{Intuition:} ER estimates representational expansiveness (active dimensionality). \\

\textbf{Participation Ratio (PR)} &
With eigenvalues $\{\lambda_i\}_{i=1}^{K}$ of $C$,
\[
PR=\frac{1}{K}\,\frac{\big(\sum_{i=1}^{K} \lambda_i\big)^2}{\sum_{i=1}^{K} \lambda_i^2}.
\]
&
\textbf{We report:} maximum PR across prefix lengths.
\newline \emph{Intuition:} PR reflects how evenly variance is distributed (higher means more diffuse). \\

\bottomrule
\end{tabular}
\caption{\textbf{Our proposed latent metrics used in LPP.} Symbols: $P_{\theta}(v \mid x)$ is the model next-token distribution over the vocabulary $\mathcal{V}$; $h_{\ell}(x;\theta)$ are layer-$\ell$ hidden states; $C=\mathrm{Cov}(h)$; $\lambda_i$ are the eigenvalues of $C$ and $K=\min(N,d)$ is their number, for $N$ pooled token representations and hidden dimension $d$. All three metrics are normalized to the unit interval (entropy by $\log|\mathcal{V}|$; ER and PR by $K$). We summarize behavior using extremal statistics (min-entropy, max-ER, max-PR). In Figure~\ref{fig:results}, all three are negatively associated with benchmark scores, although most associations are not statistically significant with eight models.}
\label{tab:metrics}
\end{table}

%% file: tables/tasks.tex
\begin{table}[!t]
\centering
\rowcolors{2}{gray!10}{white}
\begin{tabular}{p{4cm} p{9cm} p{3cm}}
\toprule
\textbf{Task} & \textbf{Input} & \textbf{Response} \\
\midrule
Ambiguous Reasoning (AR) & \texttt{Consider the ambiguous prefix and two possible senses. First, judge the prefix alone as AMBIGUOUS or NOT AMBIGUOUS. Then, after reading the hint, choose the correct option A or B. Respond strictly as: ambiguous status=AMBIGUOUS or NOT AMBIGUOUS and answer=A or B. Prefix: She deposited money at the, Options: A. bank branch or B. river bank. Hint: The muddy shore was slippery after the rain. Your response:} & \texttt{ambiguous status=AMBIGUOUS; answer=B} \\
Symbolic Pattern Completion (SPC) &
\texttt{You are given a symbolic sequence. Continue it by writing exactly the next 3 symbols, without spaces or explanations. Sequence: DYDYDYDYDYDY Answer:} & \texttt{DYD} \\
\bottomrule
\end{tabular}
\caption{\textbf{Examples from LPP-driven synthetic evaluation tasks.} This table illustrates two diagnostic task types used in our LPP framework: Ambiguous Reasoning (AR) and Symbolic Pattern Completion (SPC). These tasks are generated to probe specific latent capabilities of LLMs that are not reliably captured by conventional benchmarks. In the AR task, models must resolve lexical ambiguity through contextual inference. The prefix ``She deposited money at the'' admits two possible continuations --- ``bank branch'' (financial) and ``river bank'' (geographic). The model must use the disambiguating hint to select the intended continuation; only this final answer choice is scored. Success on AR reflects a model's ability to use disambiguating context. In contrast, the SPC task evaluates the model's ability to detect latent symbolic rules and extend them predictively. Given a structured input sequence like DYDYDYDYDYDY, the model must infer the underlying repetition and continue the sequence correctly. These SPC examples are synthetically constructed from rule-based templates (e.g., alternating patterns, mirrored sequences, or modular increments), emphasizing compression and generalization over token memorization. Together, AR and SPC stress-test models on linguistic ambiguity and symbolic abstraction, respectively. In our results, strong performance is associated with different LPP characteristics: a low entropy floor for AR, and a low participation ratio (PR) and low effective rank for SPC.}
\label{tab:examples}
\end{table}